\documentclass{article}

\usepackage{PRIMEarxiv}

\usepackage[utf8]{inputenc} 
\usepackage[T1]{fontenc}    
\usepackage{hyperref}       
\usepackage{url}            
\usepackage{booktabs}       
\usepackage{amsfonts}       
\usepackage{nicefrac}       
\usepackage{microtype}      
\usepackage{lipsum}
\usepackage{fancyhdr}       
\usepackage{graphicx}  
\usepackage{multirow}
\usepackage{wrapfig}
\usepackage[table]{xcolor}
\usepackage{colortbl}

\graphicspath{{media/}}     
\usepackage{amsmath}
\usepackage{pifont}
\usepackage{amssymb}

\title{Masked Omics Modeling for Multimodal Representation Learning across Histopathology and Molecular Profiles
}

\author{
  Lucas Robinet \\
  Oncopole Claudius Régaud \\
  IRT Saint Exupéry \\
  INSERM Cancer Research Center of Toulouse \\
  Toulouse\\
  \texttt{robinet.lucas@iuct-oncopole.fr} \\
   \And
  Ahmad Berjaoui \\
  IRT Saint Exupéry \\
  INSERM Cancer Research Center of Toulouse \\
  Toulouse\\
  \texttt{ahmad.berjaoui@irt-saintexupery.com} \\
  \And
  Elizabeth Cohen-Jonathan Moyal  \\
  Oncopole Claudius Régaud \\
  INSERM Cancer Research Center of Toulouse \\
  Toulouse\\
  \texttt{moyal.elisabeth@iuct-oncopole.fr} \\
  }
\newcommand{\morpheus}{\textsc{morpheus}}
\begin{document}

\maketitle

\begin{abstract}
Self-supervised learning (SSL) has driven major advances in computational pathology by enabling the learning of rich representations from histopathology data.
Yet, tissue analysis alone may fall short in capturing broader molecular complexity, as key complementary information resides in high-dimensional omics profiles such as transcriptomics, methylomics, and genomics.
To address this gap, we introduce \morpheus, the first multimodal pre-training strategy that integrates histopathology images and multi-omics data within a shared transformer-based architecture.
At its core, \morpheus\ relies on a novel masked omics modeling objective that encourages the model to learn meaningful cross-modal relationships.
This yields a general-purpose pre-trained encoder that can be applied to histopathology alone or in combination with any subset of omics modalities.
Beyond inference, \morpheus\ also supports flexible any-to-any omics reconstruction, enabling one or more omics profiles to be reconstructed from any modality subset that includes histopathology.
Pre-trained on a large pan-cancer cohort, \morpheus\ shows substantial improvements over supervised and SSL baselines across diverse tasks and modality combinations.
Together, these capabilities position it as a promising direction for the development of multimodal foundation models in oncology.
Code is publicly available at \href{https://github.com/Lucas-rbnt/MORPHEUS}{https://github.com/Lucas-rbnt/MORPHEUS}
\end{abstract}

\section{Introduction}
In recent years, self-supervised learning (SSL) has emerged as an effective strategy for leveraging large-scale unlabeled data to learn transferable representations for a wide range of downstream tasks \cite{chen_simple_2020,caron_emerging_2021,he_masked_2022}.
Histopathology is no exception: whole-slide images (WSIs) offer a rich view of tumor biology that can inform both diagnosis and prognosis \cite{filiot_scaling_2024}.
Because of their large size, often exceeding \( 10^5 \times 10^5\) pixels, WSIs are typically tessellated into smaller patches (\textit{e.g.,} \(256\times 256\)), naturally steering SSL efforts toward patch-level representations.
Early approaches extract patch embeddings using generic networks pre-trained on ImageNet \cite{lu_data-efficient_2021}. 
While this provides a convenient starting point, it is limited by the substantial domain gap between natural and histopathology images.
Recently, the availability of larger, well-curated pathology datasets has made it possible to develop robust pathology-specific encoders \cite{filiot_scaling_2024,vorontsov_virchow_2023,chen_towards_2024,lu_visual-language_2024}, marking a significant shift in the field.
SSL can also be extended beyond patch-level features to derive slide-level embeddings that reflect global tissue context \cite{chen_scaling_2022,jiang_masked_2024,song_morphological_2024,xu2024gigapath,jaume_transcriptomics-guided_2024}.
The resulting representation can then be used for various downstream tasks, such as survival analysis or biomarker prediction.

However, beyond morphology, a comprehensive understanding of cancer requires integrating complementary molecular signals that histopathology alone cannot fully capture.
Transcriptomic data (RNA) quantify gene expression levels, providing insight into cellular states and functional activity. 
DNA methylation (DNAm) reveals epigenetic regulation by measuring methylation levels at CpG sites, which influence gene accessibility and silencing patterns.
Copy-number variation (CNV) data reflect large-scale genomic gains and losses that drive tumor development and heterogeneity.
Each of these modalities offers a distinct and complementary view of the disease, contributing to a more comprehensive understanding of tumor biology and patient-specific trajectories.
\begin{figure}[!ht]
\centering
\includegraphics[scale=0.35]{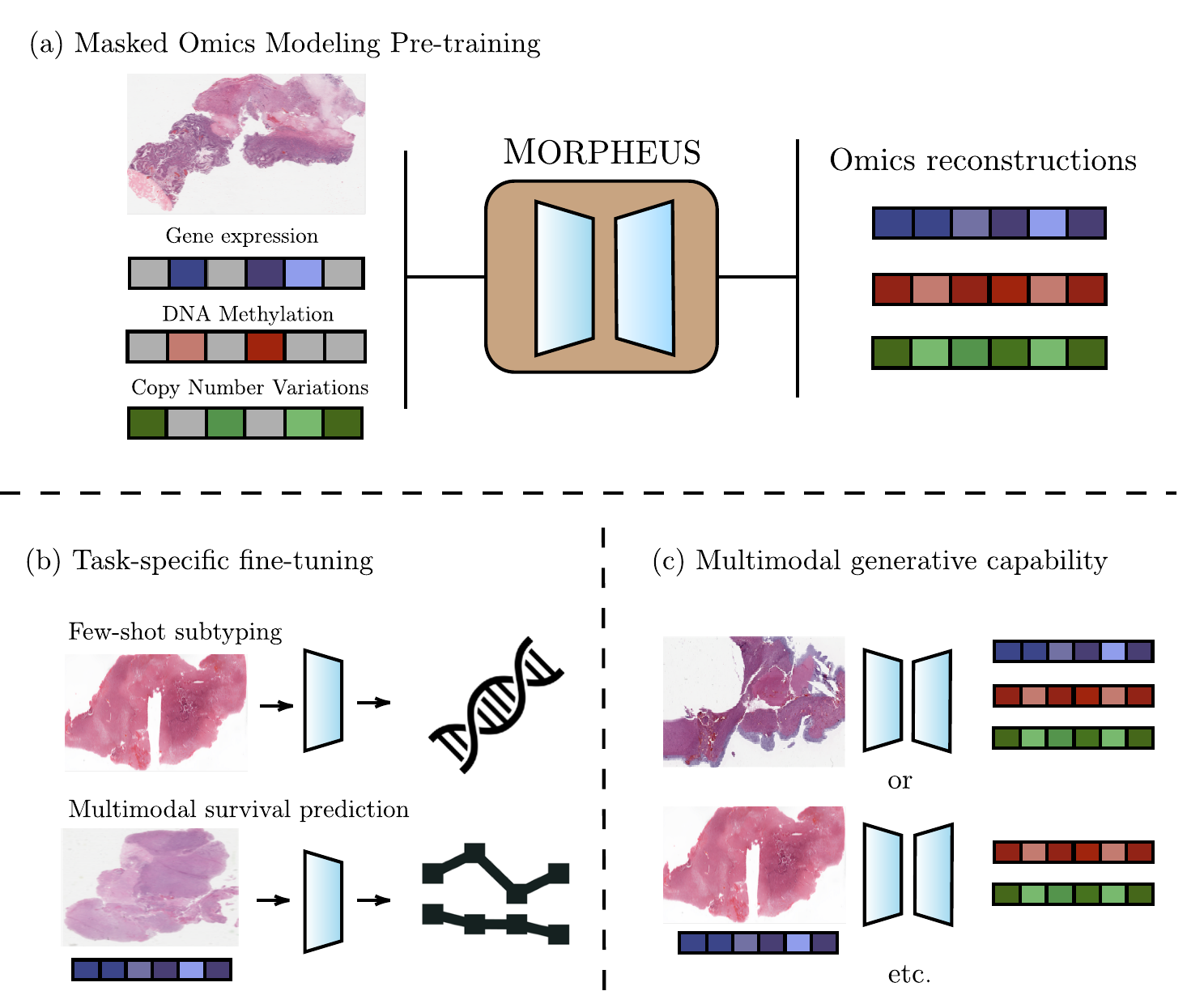}
\caption{\textbf{Overview of the proposed masked omics modeling approach.} 
(a) \morpheus\ pre-training uses histopathology and multi-omics data, learning to reconstruct masked regions of the omics profiles.
(b) The pre-trained encoder serves as a flexible multimodal backbone for downstream tasks, whether using WSI alone or in combination with omics data.
(c) \morpheus\ further enables flexible reconstruction of omics modalities from diverse modality subsets.}
\label{fig:intro-morpheus}
\end{figure}
In traditional domains such as computer vision, SSL has successfully leveraged multimodal data to build general-purpose models \cite{zamir_multimae_2022,mizrahi_4m_2023}.
In oncology, however, most multimodal integration efforts remain confined to supervised applications, predominantly survival prediction \cite{vale-silva_long-term_2021,chen_multimodal_2021,xu_multimodal_2023,jaume2023modeling,xiong_mome_2024,song2024multimodal,robinet_drim_2024}.
Such methods typically employ separate encoders for each modality and integrate them through intermediate or late fusion.
This design inherently limits their flexibility and makes them ill-suited for training generalist multimodal models.
To bridge this gap, one line of work seeks to incorporate multimodal information by aligning slide-level representations with omics embeddings through contrastive learning \cite{jaume_transcriptomics-guided_2024,vaidya2025moleculardrivenfoundationmodeloncologic,BERJAOUI2025103191}.
Yet these approaches use omics modalities only as auxiliary supervision to guide a WSI encoder, rather than treating them as full-fledged modalities.
As a result, no existing multimodal SSL method can jointly leverage histopathology and multi-omics data to train a general-purpose model that captures rich cross-modal relationships.

In this work, we introduce \morpheus\ (Masked Omics modeling for multimodal RePresentation learning across Histopathology and molEcUlar profileS), the first SSL framework specifically designed for multimodal cancer data.
\morpheus\ extends beyond histopathology by treating multi-omics profiles as full input modalities.
Concretely, we propose a simple yet powerful training paradigm: masked omics modeling.
We randomly mask portions of each omics profile and train a multimodal transformer to reconstruct the missing elements using the unmasked features together with the corresponding WSI.
To enable joint learning within a single transformer, each modality is first mapped to a set of tokens in a shared representation space.
For WSI, the high number of patches motivates the use of a Perceiver-based architecture \cite{jaegle_perceiver_2021} to aggregate redundant patch features into a compact latent representation.
Hence, rather than relying on computationally heavy patch-level embeddings, the model learns a small set of prototype tokens that capture the most salient histological patterns.
For RNA, we organize the tokens according to known pathways, grouping together genes that participate in the same biological processes.
In contrast, DNAm and CNV features are gathered based on their chromosomal positions to preserve local genomic context.
All feature groups across modalities are subsequently projected into a unified representation space.
We randomly mask a subset of omics tokens and feed the remaining omics tokens together with the WSI-derived tokens into a single shared transformer encoder.
Then, each omics modality has its own decoder to reconstruct the masked content from the multimodal encoded context.
After this pre-training stage, the resulting encoder can be seamlessly applied to diverse downstream tasks, as shown in Figure~\ref{fig:intro-morpheus}.

Our contributions can be summarized as follows: (1) we propose the first SSL framework for multimodal cancer data that jointly integrates histopathology and multi-omics through a masked modeling objective. 
To our knowledge, it is also the only approach that employs a shared encoder to process such heterogeneous biological modalities.
(2) Pre-trained on a large pan-cancer cohort, it achieves strong performance when fine-tuned on downstream tasks such as biomarker prediction and survival analysis.
(3) \morpheus\ enables flexible any-to-any omics reconstruction, allowing any target omics profile to be generated from any combination of input modalities that includes at least a WSI.

\section{Related work}
We propose to distinguish two main categories.
The first focuses on SSL applied to WSIs, a direction that has been extensively explored. 
The second concerns multimodal learning and provides valuable insights into methodological strategies for working with heterogeneous modalities.

\subsection{SSL in histopathology}
SSL in histopathology typically falls into two categories, depending on the scale of the learned representation: patch-level methods and slide-level methods.
\\
\newline
\textbf{Patch-level methods} operate on small regions of the slide, capturing fine-grained morphological patterns for local objectives such as tumor localization or nuclei detection.
Early patch-level approaches rely on contrastive learning \cite{WANG2022102559,CIGA2022100198} to pull together augmented views of the same patch while pushing apart representations of different patches.
Recently, with the growing availability of large-scale datasets, foundation model training paradigms have been adapted to histopathology to learn robust patch encoders.
Among these, DINOv2-style student-teacher frameworks have proven particularly effective: the student network learns to match representations from a momentum-updated teacher, leading to improved performance on diverse downstream tasks \cite{chen_towards_2024,vorontsov_virchow_2023}.
Finally, masked image modeling on WSI patches has also shown promise, with models learning to reconstruct missing regions from partially masked inputs \cite{filiot_scaling_2024}.
Patch-level SSL remains crucial, as it largely influences the final outcome at the slide level \cite{mammadov_self-supervision_2025}.
\\
\newline
\textbf{Slide-level methods} aim to learn global WSI representations designed for clinically relevant tasks such as prognosis and tumor subtyping.
Despite growing interest, this field is still in its early stages.
Some approaches leverage the hierarchical nature of WSI by progressively encoding multi-scale information, yielding representations that capture both local detail and global context \cite{chen_scaling_2022}. 
Another line of work, Prov-GigaPath \cite{xu2024gigapath}, adapts masked image modeling for slide-level representation learning.
Patch embeddings are first extracted by a pre-trained encoder, and a subset of these embeddings is randomly masked. 
A transformer is then trained to reconstruct the masked embeddings from the visible context, yielding a global WSI representation.
Nevertheless, deriving such embeddings remains particularly challenging as WSIs contain both highly redundant regions and a wide variety of local tissue patterns.
To address this, prototype-based learning has been introduced to summarize recurring structures into a compact set of informative prototypes \cite{song_morphological_2024}.
These prototypes retain the most essential morphological information while reducing redundancy.

\subsection{Multimodal learning}
\textbf{Supervised multimodal learning} leverages the complementarity of biological data, particularly through the integration of WSI and gene expression.
These approaches have been widely applied to survival prediction \cite{chen_multimodal_2021,chen_pathomic_2022,jaume2023modeling,song2024multimodal}.
A commonly used integration strategy is late fusion, valued for its simplicity and robustness to modality heterogeneity \cite{steyaert_multimodal_2023}. 
In this approach, each modality is first encoded separately to produce separate embeddings.
The latter are then aggregated and passed through a joint predictor to make the final decision.
However, such approaches fail to capture informative cross-modal interactions, as the modalities are only combined at the prediction stage.
To circumvent this, one can employ attention-based fusion, representing histopathology as a set of image patches and grouping gene expression features by gene families \cite{chen_multimodal_2021} or biological pathways \cite{jaume2023modeling,song2024multimodal}.
While this enables richer cross-modal interactions, it introduces a computational challenge: the large number of image patches leads to quadratic complexity in standard attention mechanisms. 
To make this tractable, one must either adopt approximate attention methods \cite{jaume2023modeling} or compress histopathology patches into a smaller set of representative tokens, often obtained through prototype-based aggregation \cite{song2024multimodal}.
Finally, some approaches have extended beyond two modalities to incorporate additional omics sources, such as DNAm or CNV \cite{vale-silva_long-term_2021,robinet_drim_2024}.
They generally adopt separate encoders per modality and combine the resulting representations using straightforward fusion schemes (\textit{e.g.,} masked attention, mean or max pooling) to cope with missing modalities.
\\
\newline
\textbf{Omics guidance} draws on omics modalities as supervisory signals for training histopathology models.
In this setting, a WSI encoder learns to align its representations with those of one or more omics encoders using a contrastive loss on paired samples \cite{jaume_transcriptomics-guided_2024,vaidya2025moleculardrivenfoundationmodeloncologic}.
This strategy enables the encoder to absorb rich information from omics profiles, transferring biological knowledge into the histopathology domain.
While highly promising, these methods focus on training a strong slide-level encoder carrying molecular context and do not consider omics as full-fledged modalities.
Consequently, they are less suitable for tasks requiring an unified multimodal patient-level representation.

\section{Method}
\begin{figure*}[t]
\centering
\includegraphics[scale=0.22]{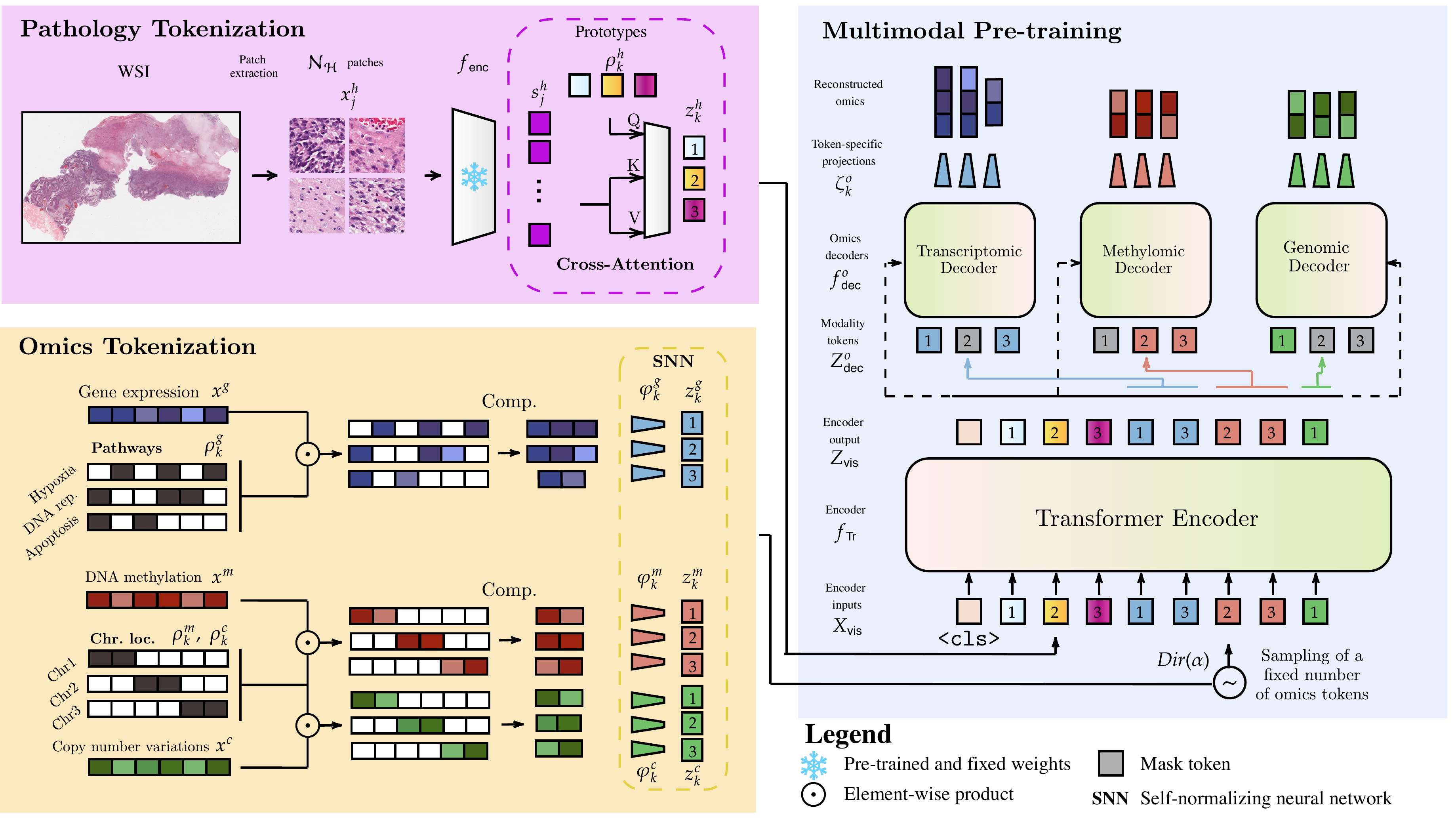}
\caption{\textbf{Overview of the pre-training framework.} 
Each modality is first mapped to a discrete set of tokens. 
WSI data are encoded using learnable prototypes through a Perceiver cross-attention mechanism \cite{jaegle_perceiver_2021}, while omics features are aggregated by functional pathways for RNA and by genomic context for DNAm and CNV. 
A proportion of omics tokens is then randomly masked, and the remaining tokens, together with all WSI tokens, are fed into a shared transformer.
After the encoding phase, mask tokens guide modality-specific decoders to reconstruct the corresponding omics profile from the multimodal representation.}
\label{fig:mtd}
\end{figure*}

The rationale behind \morpheus\ is to develop a scalable and flexible framework able to integrate a wide range of heterogeneous biological modalities, as illustrated in Figure~\ref{fig:mtd}.
In terms of notation, we define the set of omics modalities as \(\mathcal{O} = \{g,m,c\}\), referring to RNA, DNAm and CNV profiles, respectively.

\subsection{Histopathology tokenization}
Each WSI, denoted \(x^h\), is partitioned into \( \mathcal{N}_\mathcal{H} \) non-overlapping tiles at 20\(\times\) magnification, yielding a collection of image patches \( \{x^h_j\}_{j=1}^{\mathcal{N}_\mathcal{H}} \). 
The number of patches \( \mathcal{N}_\mathcal{H} \) varies across slides but typically exceeds \( 10^3\). 
A fixed, pre-trained encoder \( f_{\text{enc}} \) is then used to map each patch \( x_j^h \) into an embedding \( s_j^h = f_{\text{enc}}(x_j^h) \in \mathbb{R}^{d_\mathcal{H}} \), where \( d_\mathcal{H} \) denotes the embedding dimension.  
Using these patch-level representations directly would produce too many tokens to be efficient.
We instead adopt a Perceiver architecture \cite{jaegle_perceiver_2021} that summarizes patch-level embeddings into a fixed number of latent tokens.
Let \( \mathcal{S}^h = \{s_j^h\}_{j=1}^{\mathcal{N}_\mathcal{H}} \) denote the set of patch embeddings.
We introduce a smaller set of \(N_h\) learnable prototype vectors \( \rho_1^h, \dots, \rho_{N_h}^h \in \mathbb{R}^d \), where \( N_h << \mathcal{N}_{\mathcal{H}}\) and \(d < d_{\mathcal{H}}\) is the token dimension.
Each prototype \( \rho_k^h \) serves as a query, while the set of patch embeddings \(\mathcal{S}^h\) act as keys and values in a cross-attention mechanism:
\[z^h_k =  \text{CrossAttn}(\rho_k, \mathcal{S}^h).\]
The resulting vectors \(\{ z^h_k\}_{k=1}^{N_h} \) form a compact set of WSI tokens \(\mathcal{T}^h\).
This Perceiver-based aggregation condenses redundant yet semantically meaningful patterns across the slide into a fixed representation.

\subsection{RNA tokenization}
We similarly aim to map RNA data into a comprehensive set of tokens.
The transcriptomic profile is represented as a vector \(x^g \in \mathbb{R}^{\mathcal{N}_g}\), where \( \mathcal{N}_g \) denotes the number of genes.
To incorporate biological structure, genes are grouped into \(N_g\) pathways, each representing a set of genes involved in a common cellular process or function.
Each pathway \(k = 1, \cdots, N_g\) is associated with a binary vector \( \rho_k^g \in \{0, 1\}^{\mathcal{N}_g}\), that selects the genes belonging to that pathway.
The genes assigned to pathway \(k \) are extracted via element-wise masking and then mapped to the token space as
\[z^g_k = \varphi^g_k\big(\text{Comp.} (x^g \odot \rho_k^g)\big) \in \mathbb{R}^{d},\]
where \( \odot \) denotes element-wise multiplication and \( \text{Comp.} \) removes zero entries.
Because pathways contain different number of genes, each requires its own projection network \( \varphi^g_k \).
Specifically, \( \varphi_k^g\) is a pathway-specific self-normalizing neural network (SNN) \cite{klambauer_self-normalizing_2017} that maps the selected subset to the shared representation space.
Collecting the resulting embeddings for all pathways yields the set of transcriptomic tokens, denoted as \( \mathcal{T}^g = \{z^g_k\}_{k=1}^{N_g}\).

\subsection{DNAm and CNV tokenization}
Unlike RNA, which is structured around functional gene groupings, DNAm and CNV are more strongly influenced by genomic context. 
To better reflect this structure, we group features by genomic location.
Raw DNAm and CNV data are denoted as \( x^m \in \mathbb{R}^{\mathcal{N}_m} \) and \( x^c \in \mathbb{R}^{\mathcal{N}_c} \), with \( \mathcal{N}_m \) the number of CpG sites and \( \mathcal{N}_c \) the number of genes, respectively.
To structure these high-dimensional inputs into subsets, we define a collection of binary masks 
$\{\rho^{m}_k\}_{k=1}^{N_m}$ for DNAm and $\{\rho^{c}_k\}_{k=1}^{N_c}$ for CNV.
Then, using the same principle as for RNA, we compute a set of tokens for both DNAm and CNV
\[z^m_k  = \varphi^m_k\big(\text{Comp}(x^m \odot \rho^m_k)\big),
\]
\[
z^c_k = \varphi^c_k\big(\text{Comp}(x^c \odot \rho^c_k)\big).\]
The projection networks \( \varphi_k^m\) and \( \varphi_k^c\) are also SNN, encoding variable-sized groups into the unified representation space.
The resulting sets of DNAm and CNV tokens are denoted as \(\mathcal{T}^m = \{ z^m_k \}_{k=1}^{N_m} \) and  \( \mathcal{T}^c = \{ z^c_k \}_{k=1}^{N_c}\), respectively.

\subsection{Masking strategy}
We make the deliberate choice to mask only tokens from omics modalities, while all WSI tokens are retained.
This reflects a practical assumption aligned with typical clinical workflows: digitized tissue slides are almost always available, whereas molecular profiles may be missing.
Moreover, reconstructing WSI patches embeddings is inherently ill defined. 
Patches within the same slide can be highly diverse or contain overlapping information, making it unclear what constitutes a valid reconstruction of one patch from the others. 
Consequently, WSIs are always provided as input, while omics modalities are partially masked using a sampling strategy inspired by traditional computer vision methods \cite{mizrahi_4m_2023,zamir_multimae_2022}.
Let \(L = \sum_{o \in \mathcal{O}} |\mathcal{T}^o| \) denote the total number of omics tokens.
A global masking ratio \( r \in [0,1] \) determines the number of visible tokens to keep, \(
L_{\text{vis}} = \left\lfloor (1 - r) \cdot L \right\rfloor
\).
To allocate these tokens across modalities, we sample modality-specific weights from a Dirichlet distribution with concentration parameter \(\alpha\):
\[(w^g, w^m, w^c) \sim \text{Dir}(\alpha), \quad \text{and} \quad  w^g + w^m + w^c = 1.\]
Each weight specifies the proportion of visible tokens for modality \(o\).
The corresponding number of tokens is then drawn uniformly from \( \mathcal{T}^o \) to form the subset \( \mathcal{T}^o_{\text{vis}} \subset  \mathcal{T}^o\).
Dirichlet sampling allows a broad range of masking scenarios, where some modalities may be fully masked while others remain entirely visible, resulting in diverse cross-modal patterns.

\subsection{Multimodal encoder}
Visible tokens from all omics modalities are then concatenated with all WSI tokens to form the multimodal input sequence. 
As in ViT \cite{dosovitskiy_image_2021}, we prepend a \texttt{<cls>} token that aggregates information from all tokens into a global representation.
For notational simplicity, we refer to this operation using set notation, although it corresponds in practice to the concatenation of all individual tokens.
This yields the following flattened sequence:
\[X_{\text{vis}} = \big[\texttt{<cls>}, \mathcal{T}^h, \mathcal{T}^g_{\text{vis}}, \mathcal{T}^m_{\text{vis}}, \mathcal{T}^c_{\text{vis}} \big] \in \mathbb{R}^{(N_h + L_{\text{vis}} + 1) \times d}.\] 
The latter is subsequently fed into a standard transformer encoder \(f_{\text{Tr}}\)
\[
Z_{\text{vis}} = f_{\text{Tr}}(X_{\text{vis}}) \in \mathbb{R}^{(N_h + L_{\text{vis}} + 1) \times d}.
\]

\subsection{Omics decoders}
We employ modality-specific decoders: a transcriptomic decoder for RNA, a methylomic decoder for DNAm, and a genomic decoder for CNV, all sharing the same architecture.
First, the multimodal representation \(Z_{\text{vis}}\) is linearly projected into a specific representation space.
As illustrated in Figure~\ref{fig:mtd}, the decoder input sequence \( Z^o_{\text{dec}} \in \mathbb{R}^{|\mathcal{T}^o| \times d} \) is then obtained by inserting mask tokens at the missing positions within the encoded token sequence for modality \(o\).
A learnable position encoding is also added to each token to indicate its corresponding group,
Without this, the decoder would be unable to distinguish between different mask tokens.
For each modality, we use a transformer-based decoder \(f^o_{\text{dec}}\).
Each block begins with cross-attention, allowing the modality-specific sequence \( Z^o_{\text{dec}} \) to attend to the multimodal context \(Z_{\text{vis}}\).
This is then followed by a self-attention layer
Formally, the decoded embeddings are obtained through
\[Z^o_k = f^o_{\text{dec}} (Z^o_{\text{dec}}, Z_{\text{vis}})_k,\]
where \(k\) indexes the token corresponding to group \(k\).
To account for differences in original sizes, each decoded tokens \(Z^o_k\) is processed through a group-specific projection \(\zeta_k^o\) which maps it back to its original feature space.
\morpheus\ is trained with a mean absolute error loss computed only on masked tokens.

\section{Experiments and results}
To provide the most rigorous evaluation, we benchmark \morpheus\ on a set of well-established downstream tasks.
Specifically, we evaluate:
(1) biomarker prediction from WSI alone, in both standard classification and few-shot settings; 
(2) survival analysis under different modality configurations (WSI, WSI+RNA, and all modalities), with comprehensive details provided in Appendix~\ref{app:survival}; 
and (3) generative reconstruction, assessing the ability to recover missing omics modalities from varied input combinations.

\subsection{Experimental setup}
\paragraph{Datasets.}
We conduct our experiments on the TCGA cohort \cite{weinstein2013cancer}.
Pre-training is performed on samples with all four modalities available (WSI, RNA, DNAm, and CNV), resulting in a total of 4,718 cases spanning 32 cancer types.
To avoid any data leakage, all downstream evaluations are performed on patients excluded from this pre-training cohort.
For survival prediction, we retain only patients with all four modalities available to ensure fair comparisons across input configurations.
For classification tasks, we include all patients with WSI available.
Full cohort breakdowns for both pre-training and downstream evaluations are provided in Appendix~\ref{app:cohorts}.

\paragraph{Preprocessing.}
WSI are partitioned into non-overlapping patches at \(20 \times\) magnification. 
Patch embeddings are obtained using UNIv2 \cite{chen_towards_2024}.
Following prior work \cite{song_morphological_2024}, we set the number of WSI prototypes to \(N_h = 16\). 
Normalized bulk RNA data are obtained from the UCSC Xena platform \cite{goldman_visualizing_2020}.
Genes are grouped into \(N_g = 50\) biological pathways based on the Hallmark gene sets from the Molecular Signatures Database (MSigDB) \cite{msigdb}.
For DNAm and CNV, we select high-variance features and group them by genomic location into 50 clusters per modality (\(N_m=50\), \(N_c = 50\)).
Additional details on normalization, preprocessing, and experimental settings are provided in Appendix~\ref{app:preprocessing}.

\paragraph{Baselines.} We compare \morpheus\ against several baselines under different input configurations.
\newline
\textbf{WSI-only.} We consider Attention-based MIL (ABMIL) \cite{ilse_attention-based_2018}, DeepSets \cite{zaheer_deep_2017}, Transformer-based MIL (TransMIL) \cite{shao_transmil_2021} and \(\textsc{Tangle}^{\dagger}\) \cite{jaume_transcriptomics-guided_2024}.
Note that we do not directly use the original \(\textsc{Tangle}\) model, as it relies on a different patch encoder.
Instead, to ensure a fair comparison, we use the same training procedure on UNIv2 embeddings, and refer to this version as \(\textsc{Tangle}^{\dagger}\).
See Appendix~\ref{app:tangle} for further details.
\newline
\textbf{WSI+RNA.} We compare against Multimodal Co-Attention Transformer (MCAT) \cite{chen_multimodal_2021},  Multimodal Optimal Transport-based Co-Attention Transformer (MOTCat) \cite{Xu_2023_ICCV}, SurvPath \cite{jaume2023modeling} and Mixture of Multimodal Experts (MoME) \cite{xiong_mome_2024}.
\newline
\textbf{All modalities.} 
We evaluate modality-specific encoders combined with standard fusion strategies, including mean, max, and concatenation fusion, following \cite{vale-silva_long-term_2021}.
Architectural details for each modality are provided in Appendix~\ref{app:tangle}.
\newline
\morpheus\ \textbf{variants.}
Across all input configurations, we evaluate two model variants: a single block transformer in both the encoder and decoders (\morpheus) and a two block variant with two layers per component (\morpheus-2L).

\paragraph{Implementation details.}
Pre-training is performed for 100 epochs with a batch size of 128. 
The mask ratio value follows common practice in computer vision \cite{he_masked_2022,mizrahi_4m_2023} and is set to \(r=0.75 \).
During both pre-training and downstream tasks, we uniformly sample 1,024 patch embeddings per WSI to maintain a fixed input size.
If multiple slides are available for a given patient, embeddings from all slides are stacked before sampling.
At inference, all patches from all WSI belonging to a given sample are used.
Experiments are conducted on a single NVIDIA GeForce RTX 3090 Ti.
The complete lists of hyperparameters used for pre-training and downstream tasks are provided in Appendix~\ref{app:hparams}.
Results are reported as mean \(\pm\) standard deviation. 
In all tables, the best result is shown in \textbf{bold}, and the second-best is \underline{underlined}.

\subsection{Downstream biomarker prediction}
We evaluate \morpheus\ across five subtyping tasks spanning diffuse glioma, breast cancer, and lung cancer.
For diffuse glioma, we consider three molecular biomarkers central to the WHO 2021 classification \cite{louis_2021_2021}: (1) IDH mutation status, distinguishing IDH-wild-type from IDH-mutant tumors; (2) ATRX mutation, associated with astrocytic lineage; and (3) 1p19q codeletion, the defining alteration of oligodendrogliomas.
To assess broader applicability beyond glioma, we additionally consider
(4) breast cancer, distinguishing invasive ductal carcinoma (IDC) from invasive lobular carcinoma (ILC); and
(5) lung cancer, distinguishing lung adenocarcinoma (LUAD) from lung squamous cell carcinoma (LUSC).
All five tasks are performed exclusively on WSI data and are evaluated under both traditional classification and few-shot subtyping settings. 
Task-specific sample counts and class distributions are provided in Appendix~\ref{app:cohorts}.

\paragraph{Regular classification.}
\renewcommand{\arraystretch}{1.2}
\tabcolsep=0.2cm
\begin{table}[t]
\caption{\textbf{Subtyping prediction performance.}
Results are reported as AUC (\%) across 5-fold cross-validation, under the WSI-only configuration.}
\label{tab:reg-sub}
\resizebox{\textwidth}{!}{
\begin{tabular}{l|cccccc}
\hline
            & \textbf{Brain} \text{\scriptsize (IDH)}              & \textbf{Brain} \text{\scriptsize (ATRX)}            & \textbf{Brain} \text{\scriptsize (1p19q)}            & \textbf{Breast}     & \textbf{Lung}     & \textbf{Overall}                 \\ \hline
DeepSets    & 91.5 $\pm$ 2.4          & 69.7 $\pm$ 6.1          & 84.9 $\pm$ 8.9          & 85.3 $\pm$ 6.0          & 87.6 $\pm$ 4.0          & 83.8 $\pm$ 7.4          \\
ABMIL       & 93.1 $\pm$ 2.7          & 65.3 $\pm$ 6.9          & 75.6 $\pm$ 4.3          & 61.3 $\pm$ 8.4          & 92.1 $\pm$ 3.7          & 77.5 $\pm$ 13.2         \\
TransMIL    & 94.1 $\pm$ 1.9          & 73.9 $\pm$ 4.2          & 89.1 $\pm$ 2.9          & 84.2 $\pm$ 6.0          & 95.8 $\pm$ 2.8          & 87.4 $\pm$ 7.9          \\
\textsc{Tangle}$^\dagger$      & 95.6 $\pm$ 1.9          & 74.7 $\pm$ 6.4          & 82.1 $\pm$ 6.9          & 87.1 $\pm$ 6.2          & 97.0 $\pm$ 2.1          & 87.3 $\pm$ 8.4          \\
\rowcolor{gray!15}
\morpheus\    & \textbf{97.3 $\pm$ 1.1} & \underline{81.8 $\pm$ 3.7}          & \textbf{95.7 $\pm$ 2.2} & \underline{92.1 $\pm$ 2.4}          & \textbf{97.5 $\pm$ 1.5} & \textbf{92.9 $\pm$ 5.9} \\
\rowcolor{gray!15}
\morpheus-2L & \underline{96.5 $\pm$ 1.1}          & \textbf{82.6 $\pm$ 3.7} & \underline{95.1 $\pm$ 2.2}          & \textbf{92.5 $\pm$ 1.3} & \underline{97.4 $\pm$ 1.7}          & \underline{92.8 $\pm$ 5.4}          \\ \hline
\end{tabular}}
\end{table}
In this setting, we use the full held-out cohort and evaluate all models using 5-fold cross-validation.
Training hyperparameters are listed in Appendix~\ref{app:hparams}.
Table~\ref{tab:reg-sub} reports the subtyping prediction performance. 
Both \morpheus\ variants achieve the top two scores across all five tasks.
On average, \morpheus\ yields a 6.3\% relative improvement over the strongest baseline.
The largest gains are observed on the more challenging tasks with lower overall performance, such as ATRX, where a 9.5\% improvement is achieved.
These findings underscore the effectiveness of our pre-training strategy in transferring multimodal knowledge to a WSI-only input configuration.

\paragraph{Few-shot classification.}
\tabcolsep=0.12cm
\begin{table}[ht]
\caption{\textbf{Few‑shot brain tumor subtyping prediction performance.} 
Results are reported as AUC (\%) for different \(k\)-shot scenarios, averaged over 10 random samplings.}
\label{tab:fs_results_brain}
\resizebox{\textwidth}{!}{
\begin{tabular}{lccc|ccc|ccc}
\hline
            & \multicolumn{3}{c|}{\textbf{ATRX}}                                           & \multicolumn{3}{c|}{\textbf{IDH}}                                           & \multicolumn{3}{c}{\textbf{1p19q}}                                            \\
            & \(k=1\)                  & \(k=5\)                 & \(k=10\)                & \(k=1\)                 & \(k=5\)                 & \(k=10\)                & \(k=1\)                  & \(k=5\)                 & \(k=10\)                \\ \hline
DeepSets   & 51.2 $\pm$ 3.4           & 59.1 $\pm$ 5.8          & 62.4 $\pm$ 6.2          & 59.1 $\pm$ 7.3          & 74.2 $\pm$ 6.7          & 76.3 $\pm$ 6.5          & 56.3 $\pm$ 7.9           & 65.6 $\pm$ 6.9          & 74.2 $\pm$ 4.3          \\
ABMIL     & 56.9 $\pm$ 8.0           & 63.4 $\pm$ 8.9          & 66.0 $\pm$ 6.5          & 68.4 $\pm$ 10.7         & 84.6 $\pm$ 2.9          & 86.5 $\pm$ 3.5          & 57.6 $\pm$ 13.2          & 72.7 $\pm$ 5.8          & 81.3 $\pm$ 3.5          \\
TransMIL    & 54.5 $\pm$ 7.2           & 61.8 $\pm$ 7.0          & 65.3 $\pm$ 6.2          & 63.4 $\pm$ 10.0         & 84.4 $\pm$ 3.3          & 86.2 $\pm$ 4.4          & 57.5 $\pm$ 9.8           & 73.0 $\pm$ 3.3          & 82.8 $\pm$ 3.2          \\
\textsc{Tangle}$^\dagger$     & 56.6 $\pm$ 11.7          & 63.8 $\pm$ 6.3          & 67.4 $\pm$ 6.6          & 64.8 $\pm$ 11.8         & 88.2 $\pm$ 4.6          & 91.1 $\pm$ 2.4          & \underline{60.5 $\pm$ 10.6}          & 75.5 $\pm$ 6.3          & 85.8 $\pm$ 3.0          \\
\rowcolor{gray!15}
\morpheus\     & \textbf{62.4 $\pm$ 10.0} & \textbf{72.6 $\pm$ 7.0} & \textbf{73.3 $\pm$ 6.9} & \textbf{87.8 $\pm$ 6.4} & \textbf{95.2 $\pm$ 1.2} & \textbf{96.0 $\pm$ 0.6} & 55.2 $\pm$ 20.4          & \underline{86.9 $\pm$ 4.9}          & \textbf{92.7 $\pm$ 2.5} \\
\rowcolor{gray!15}
\morpheus-2L & \underline{59.5 $\pm$ 9.6}           & \underline{70.4 $\pm$ 7.6}          & \underline{72.2 $\pm$ 5.8}         & \underline{75.9 $\pm$ 14.5}         & \underline{93.1 $\pm$ 1.8}          & \underline{94.1 $\pm$ 1.3}          & \textbf{64.0 $\pm$ 22.5} & \textbf{88.1 $\pm$ 5.0} & \underline{92.4 $\pm$ 1.5}          \\ \hline
\end{tabular}
}
\end{table}
\begin{table}[ht]
\caption{\textbf{Few‑shot breast and lung tumor subtyping prediction performance.} 
Results are reported as AUC (\%) for different \(k\)-shot scenarios, averaged over 10 random samplings.}
\small
\label{tab:fs_results_bl}
\resizebox{\textwidth}{!}{
\begin{tabular}{lccc|ccc}
\hline
            & \multicolumn{3}{c|}{\textbf{Breast}}                                         & \multicolumn{3}{c}{\textbf{Lung}}                                            \\
            & \(k=1\)                  & \(k=5\)                 & \(k=10\)                & \(k=1\)                  & \(k=5\)                 & \(k=10\)                \\ \hline
DeepSets    & 50.7 $\pm$ 6.0           & 57.9 $\pm$ 6.0          & 66.2 $\pm$ 5.0          & 53.3 $\pm$ 7.6           & 68.9 $\pm$ 3.3          & 74.9 $\pm$ 4.3          \\
ABMIL       & 51.8 $\pm$ 6.1           & 60.8 $\pm$ 3.9          & 69.1 $\pm$ 4.6          & 58.6 $\pm$ 13.7          & 75.7 $\pm$ 5.0          & 80.5 $\pm$ 4.6          \\
TransMIL    & 52.5 $\pm$ 5.6           & 61.3 $\pm$ 4.2          & 70.6 $\pm$ 4.4          & 60.1 $\pm$ 10.3          & 75.3 $\pm$ 75.7         & 83.0 $\pm$ 4.3          \\
\textsc{Tangle}$^\dagger$      & \underline{56.7 $\pm$ 15.6}          & 77.0 $\pm$ 8.3          & 85.0.$\pm$ 4.3          & 65.5 $\pm$ 19.8          & 86.8 $\pm$ 7.2          & 93.7 $\pm$ 2.2          \\
\rowcolor{gray!15}
\morpheus\    & 55.5 $\pm$ 16.2          & \underline{81.8 $\pm$ 6.7}          & \underline{87.3 $\pm$ 3.1}          & \underline{65.6 $\pm$ 16.1}          & \underline{91.4 $\pm$ 2.9}          & \textbf{96.1 $\pm$ 0.7} \\
\rowcolor{gray!15}
\morpheus-2L & \textbf{57.5 $\pm$ 15.0} & \textbf{82.2 $\pm$ 9.2} & \textbf{87.5 $\pm$ 2.9} & \textbf{74.4 $\pm$ 12.1} & \textbf{93.2 $\pm$ 1.3} & \underline{95.6 $\pm$ 0.7}          \\ \hline
\end{tabular}}
\end{table}
We further evaluate \morpheus\ in several \(k\)-shot scenarios to assess generalization under limited supervision.
We perform 10 independent runs, randomly selecting \(k\) samples per class for training and evaluating on all remaining samples.
Baselines should be compared only within the same \(k\)-shot scenario and subtyping task, as comparisons across different \(k\) are not meaningful due to differing test sets.
All hyperparameter details can be found in Appendix~\ref{app:hparams}.
Table~\ref{tab:fs_results_brain} and Table~\ref{tab:fs_results_bl} summarize few-shot performance across tasks and scenarios, with our method consistently achieving strong results across all use cases.
Overall, both \morpheus\ variants outperform all baselines, ranking first and second in the vast majority of scenarios and tasks.
The only exceptions occur in the (\(k=1\)) scenario for 1p19q prediction and for breast cancer subtyping, where \(\textsc{Tangle}^{\dagger}\) ranks second and \morpheus\ falls to third place.
In the (\(k=10\)) scenario for brain subtyping tasks, \morpheus\ achieves substantial gains over the best supervised baselines, improving AUC by 11.1\%, 11.0\%, and 12.0\% on ATRX, IDH, and 1p19q, respectively. 
It also surpasses \(\textsc{Tangle}^{\dagger}\), the strongest pre-training alternative, by 8.8\%, 5.4\%, and 8.0\% on these same tasks.

\subsection{Downstream survival analysis}
We further evaluate \morpheus\ on patient prognosis prediction.
Concretely, we consider three fine-tuning configurations: (1) a WSI-only setting, (2) a WSI+RNA setting, and (3) a fully multimodal setting.
More details on the cohort for this task and training hyperparameters are provided in Appendix~\ref{app:cohorts} and Appendix~\ref{app:hparams}, respectively.
All models are evaluated using 5-fold cross-validation.
Results in Table~\ref{tab:surv} show that both variants of our method perform consistently across input configurations.
On the LUAD cohort, however, \morpheus\ exhibits markedly lower and highly variable survival performance, particularly when using full multimodal inputs, which drives the lower overall mean observed for this cohort.
This behavior is not unexpected in survival analysis under limited-data conditions: after splitting cohorts between pre-training and fine-tuning, the resulting survival sets are small, leading to unstable concordance-based metrics with high variance \cite{beca_impact_2021,riley_stability_2023}.
This instability is particularly emphasized in \(\textsc{Tangle}^{\dagger}\), the closest pre-training baseline in the WSI-only setting.
Consequently, all methods show only modest performance, often close to random, underscoring the difficulty of evaluating survival models in such limited-data settings.
A clearer picture emerges when examining the GBMLGG cohort, where metrics are consistently higher across considered methods. 
On this cohort, we observe that \morpheus\ performance improves as additional modalities are provided, highlighting the benefits of the proposed multimodal fusion.
\begin{table}[t]
\caption{\textbf{Survival prognosis performance.}
Results are reported as concordance index across 5-fold cross-validation, under different input configurations.}
\label{tab:surv}
\resizebox{\textwidth}{!}{
\begin{tabular}{clllllcl}
\hline
\multicolumn{8}{c}{\textbf{C-Index ($\uparrow$)}}                                                                                                                                                                                                                                                                                                       \\
\multicolumn{1}{l}{}                                                    & \multicolumn{1}{c}{} & \textbf{BRCA}                     & \multicolumn{1}{c}{\textbf{GBMLGG}}                  & \multicolumn{1}{c}{\textbf{COADREAD}}                 & \multicolumn{1}{c}{\textbf{STAD}}                    & \textbf{LUAD}                    & \multicolumn{1}{c}{\textbf{Overall}}                  \\ \hline
\multirow{6}{*}{WSI}                                                    & DeepSets             & 53.1 $\pm$ 8.1           & \multicolumn{1}{c}{58.4 $\pm$ 4.4}          & \multicolumn{1}{c}{47.4 $\pm$ 29.2}          & \multicolumn{1}{c}{49.5 $\pm$ 14.9}         & 45.6 $\pm$ 7.5          & \multicolumn{1}{c}{50.8 $\pm$ 4.5}           \\
                                                                        & ABMIL                & \underline{57.0 $\pm$ 6.2}           & \multicolumn{1}{c}{78.9 $\pm$ 5.7}          & \multicolumn{1}{c}{50.2 $\pm$ 19.6}          & \multicolumn{1}{c}{50.3 $\pm$ 9.4}          & \underline{58.4 $\pm$ 4.3}          & \multicolumn{1}{c}{59.0 $\pm$ 10.5}          \\
                                                                        & TransMIL             & 54.2 $\pm$ 13.2          & \multicolumn{1}{c}{\underline{81.4 $\pm$ 3.6}}          & \multicolumn{1}{c}{43.9 $\pm$ 11.9}          & \multicolumn{1}{c}{52.4 $\pm$ 11.0}         & 55.1 $\pm$ 8.6          & \multicolumn{1}{c}{57.4 $\pm$ 12.6}          \\
                                                                        & \textsc{Tangle}$^\dagger$               & 50.2 $\pm$ 11.6          & \multicolumn{1}{c}{76.3 $\pm$ 3.2}          & \multicolumn{1}{c}{50.6 $\pm$ 14.2}          & \multicolumn{1}{c}{50.9 $\pm$ 8.8}          & \textbf{59.4 $\pm$ 4.1} & \multicolumn{1}{c}{57.5 $\pm$ 10.0}          \\
                                                                        & \cellcolor{gray!15}\morpheus\             & \cellcolor{gray!15}\textbf{58.2 $\pm$ 10.4} & \cellcolor{gray!15}\textbf{81.5 $\pm$ 4.1} & \cellcolor{gray!15}\textbf{59.0 $\pm$ 16.8} & \cellcolor{gray!15}\textbf{54.9 $\pm$ 5.0} & \cellcolor{gray!15}52.2 $\pm$ 8.4          & \cellcolor{gray!15}\textbf{61.2 $\pm$ 10.5} \\
                                                                        & \cellcolor{gray!15}\morpheus-2L          & \cellcolor{gray!15}56.7 $\pm$ 13.3          & \cellcolor{gray!15}78.4 $\pm$ 3.6          & \cellcolor{gray!15}\underline{56.8 $\pm$ 10.6}          & \cellcolor{gray!15}\underline{54.1 $\pm$ 14.0}         & \cellcolor{gray!15}55.5 $\pm$ 2.4          & \cellcolor{gray!15}\underline{60.3 $\pm$ 9.1 }          \\ \hline
\multirow{6}{*}{\begin{tabular}[c]{@{}c@{}}WSI \\ +\\ RNA\end{tabular}} & MCAT                 & \textbf{59.9 $\pm$ 11.5}          & 80.7 $\pm$ 2.8                              & 51.7 $\pm$ 15.2                              & 53.6 $\pm$ 10.9                             & 53.4 $\pm$ 6.1          & 59.9 $\pm$ 10.8                              \\
                                                                        & MOTCat               & 54.8 $\pm$ 11.3          & 83.7 $\pm$ 4.1                              & 55.7 $\pm$ 15.7                              & 50.1 $\pm$ 10.1                             & 51.3 $\pm$ 8.3          & 59.1 $\pm$ 12.5                              \\
                                                                        & SurvPath             & 50.3 $\pm$ 13.3          & 79.6 $\pm$ 4.5                              & 44.2 $\pm$ 10.9                              & 55.8 $\pm$ 8.3                              & 55.9 $\pm$ 6.7          & 57.1 $\pm$ 12.0                              \\
                                                                        & MoME                 & 47.6 $\pm$ 19.1          & 74.1 $\pm$ 11.7                             & 45.0 $\pm$ 25.1                              & 49.7 $\pm$ 11.5                             & 48.0 $\pm$ 10.0         & 52.9 $\pm$ 10.7                              \\
                                                                        & \cellcolor{gray!15}\morpheus\             & \cellcolor{gray!15}55.0 $\pm$ 12.2          & \cellcolor{gray!15}\textbf{84.9 $\pm$ 4.5}                     & \cellcolor{gray!15}\underline{62.3 $\pm$ 15.6}                              & \cellcolor{gray!15}\textbf{59.5 $\pm$ 5.5}                     & \cellcolor{gray!15}\textbf{56.3 $\pm$ 4.3} & \cellcolor{gray!15}\underline{63.6 $\pm$ 10.9}                              \\
                                                                        & \cellcolor{gray!15}\morpheus-2L          & \cellcolor{gray!15}\underline{56.7 $\pm$ 19.3} & \cellcolor{gray!15}\underline{84.6 $\pm$ 3.1}                              & \cellcolor{gray!15}\textbf{64.5 $\pm$ 18.6}                     & \cellcolor{gray!15}\underline{56.4 $\pm$ 7.6}                              & \cellcolor{gray!15}\underline{56.2 $\pm$ 5.2}          & \cellcolor{gray!15}\textbf{63.7 $\pm$ 10.9}                     \\ \hline
\multirow{5}{*}{All modalities}                                         & Mean                 & 57.3 $\pm$ 5.3           & 83.9 $\pm$ 3.4                              & 50.8 $\pm$ 11.0                              & 51.0 $\pm$ 10.5                             & \underline{58.2 $\pm$ 2.1}          & 60.2 $\pm$ 12.2                              \\
                                                                        & Max                  & 57.0 $\pm$ 9.8           & 83.7 $\pm$ 2.9                              & 46.5 $\pm$ 17.1                              & 51.5 $\pm$ 2.8                              & 53.2 $\pm$ 7.3          & 58.4 $\pm$ 13.1                              \\
                                                                        & Concat              & \underline{63.9 $\pm$ 10.8}          & 84.1 $\pm$ 1.9                              & 45.1 $\pm$ 7.7                               & 54.0 $\pm$ 11.5                             & \textbf{60.8 $\pm$ 2.8}          & 61.6 $\pm$ 13.0                              \\
                                                                        & \cellcolor{gray!15}\morpheus\             & \cellcolor{gray!15}58.1 $\pm$ 18.5          & \cellcolor{gray!15}\textbf{85.8 $\pm$ 2.0}                     & \cellcolor{gray!15}\underline{54.7 $\pm$ 16.2}                              & \cellcolor{gray!15}\textbf{59.1 $\pm$ 10.0}                    & \cellcolor{gray!15}52.2 $\pm$ 9.8          & \cellcolor{gray!15}\underline{61.8 $\pm$ 11.8}                              \\
                                                                        & \cellcolor{gray!15}\morpheus-2L          & \cellcolor{gray!15}\textbf{70.6 $\pm$ 15.5} & \cellcolor{gray!15}\underline{85.0 $\pm$ 2.9}                             & \cellcolor{gray!15}\textbf{63.2 $\pm$ 14.9}                     & \cellcolor{gray!15}\underline{57.4 $\pm$ 8.0}                              & \cellcolor{gray!15}56.9 $\pm$ 8.1          & \cellcolor{gray!15}\textbf{66.8 $\pm$ 10.7}                     \\ \hline
\end{tabular}
}
\end{table}
\paragraph{Sensitivity analysis.}
A key strength of our approach lies in its architectural simplicity: \morpheus\ builds on a standard transformer architecture adapted to cancer data, without introducing task-specific modules or modality-dependent branches.
Results from the two model variants considered in our experiments indicate that performance remains stable across different encoder depths, suggesting limited sensitivity to model depth within the explored range.
We additionally examine the impact of the number of pre-training epochs on downstream performance.
A detailed analysis of the effect of increased pre-training duration on few-shot classification performance is provided in Appendix~\ref{app:sensitivity}.

\subsection{Reconstructions of omics modalities}
High-dimensional omics modalities remains costly and time-consuming to acquire, often requiring specialized equipment and extensive preprocessing.
Being able to accurately reconstruct these modalities is therefore valuable as it enables richer patient profiling and supports more informed personalized treatment decisions.
In prior work, reconstruction tasks have typically adopted a direct approach, where WSI are provided as input to a neural network trained to reconstruct a target omics profile, such as RNA \cite{schmauch_deep_2020} or DNAm \cite{hoang_prediction_2024}.
While convenient, these approaches lack multimodal flexibility, as they are restricted to a fixed source-target mapping.
\begin{figure*}[t]
\centering
\includegraphics[width=0.32\textwidth]{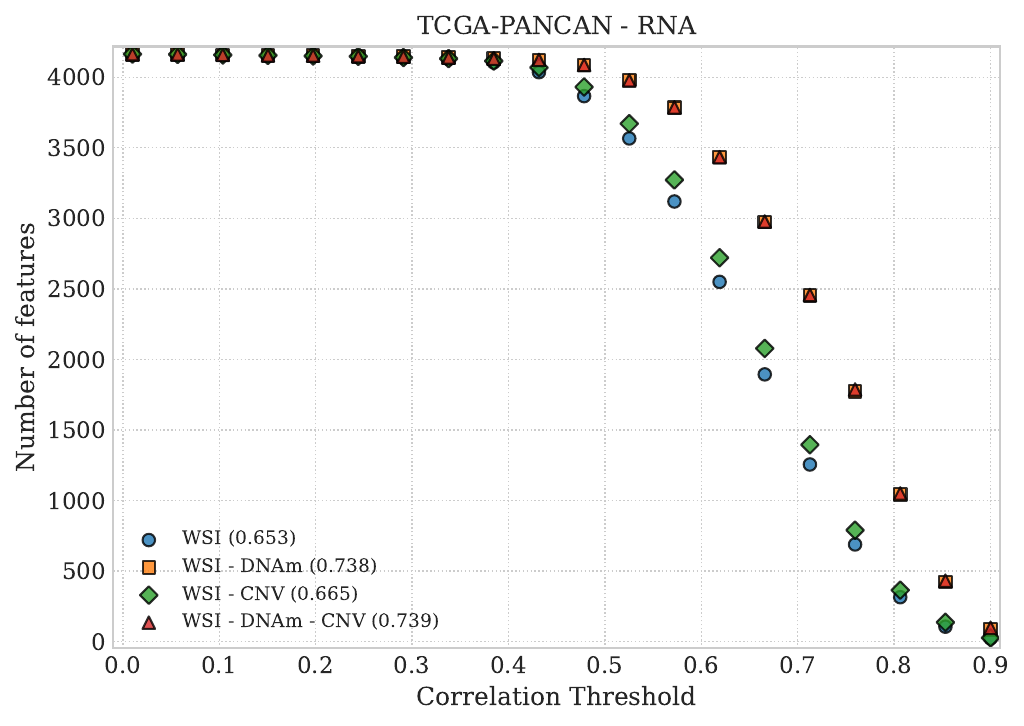}
\includegraphics[width=0.32\textwidth]{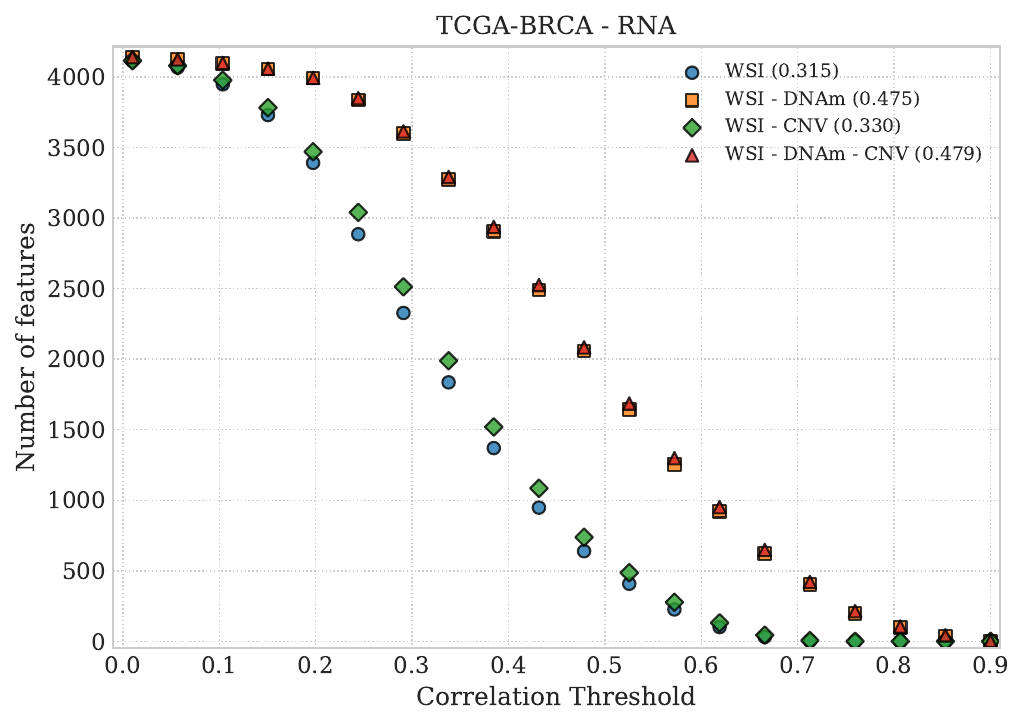}
\includegraphics[width=0.32\textwidth]{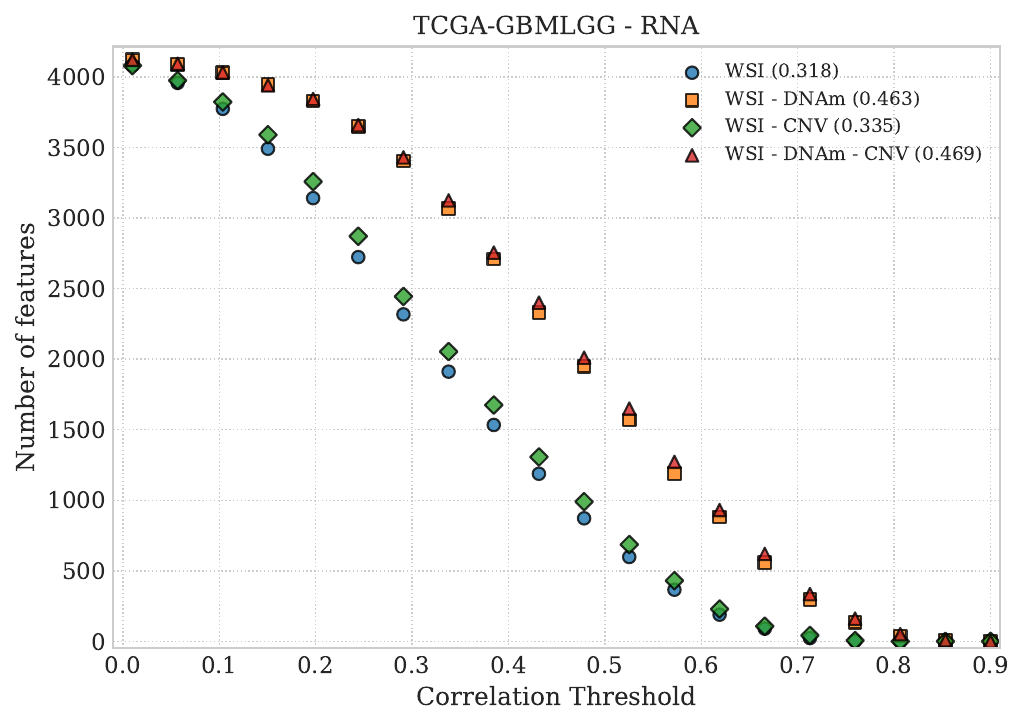} \\[0.5em]

\includegraphics[width=0.32\textwidth]{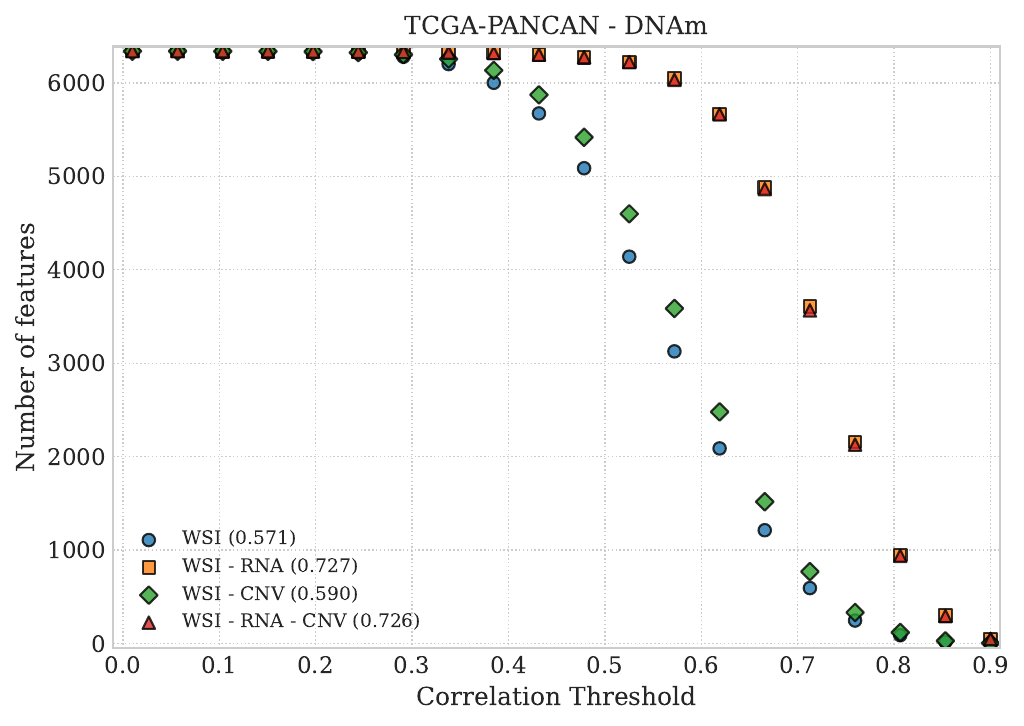}
\includegraphics[width=0.32\textwidth]{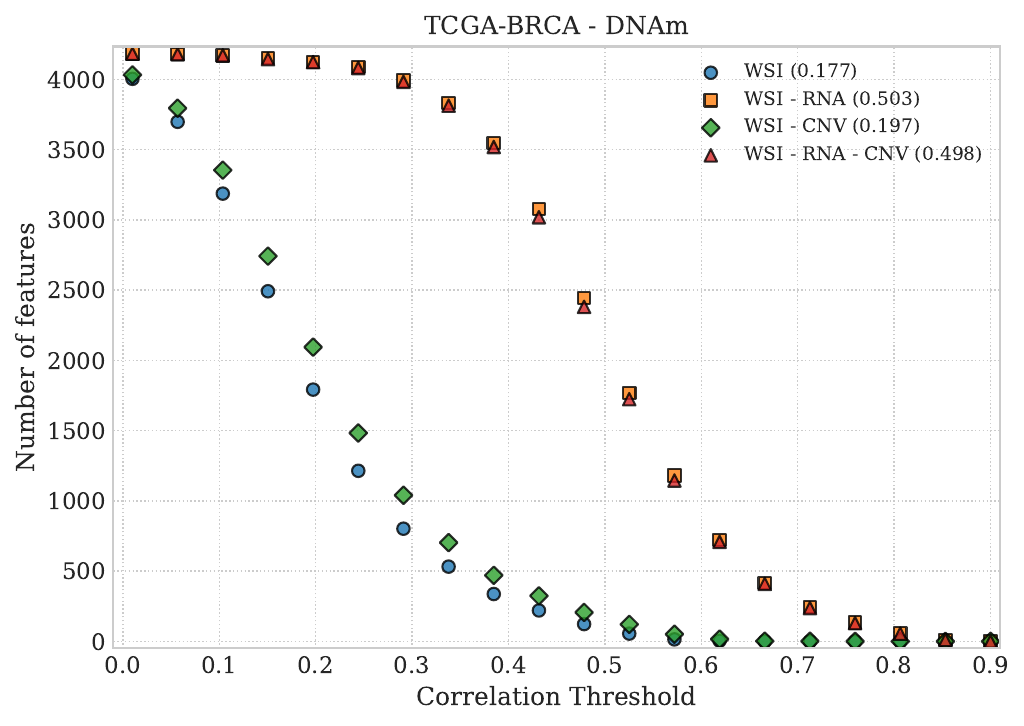}
\includegraphics[width=0.32\textwidth]{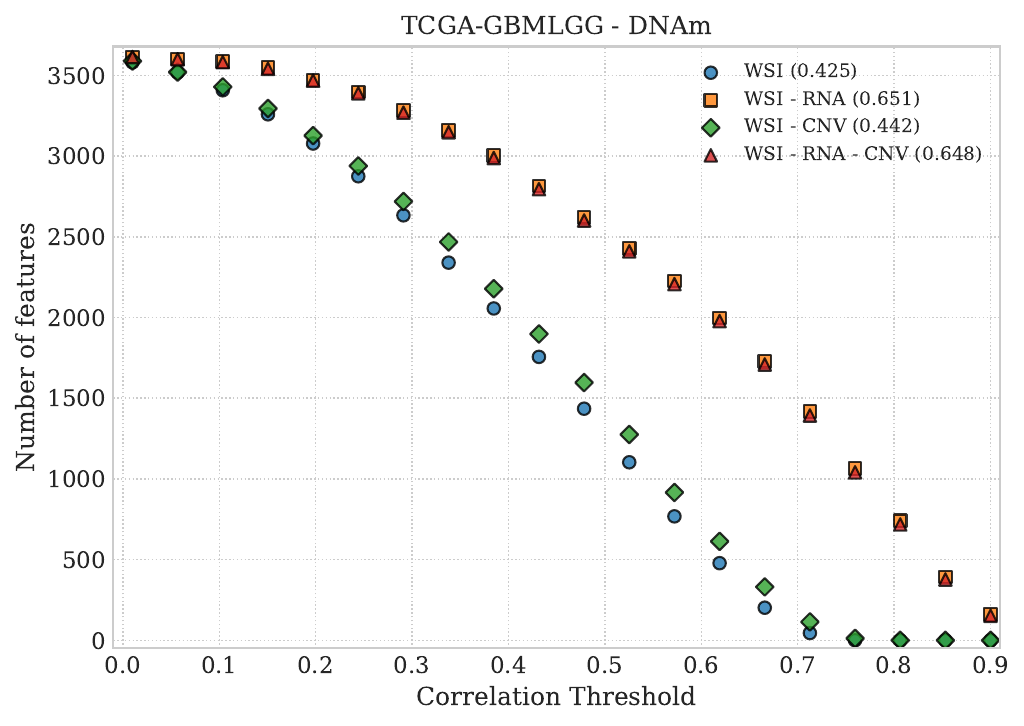}\\[0.5em]

\includegraphics[width=0.32\textwidth]{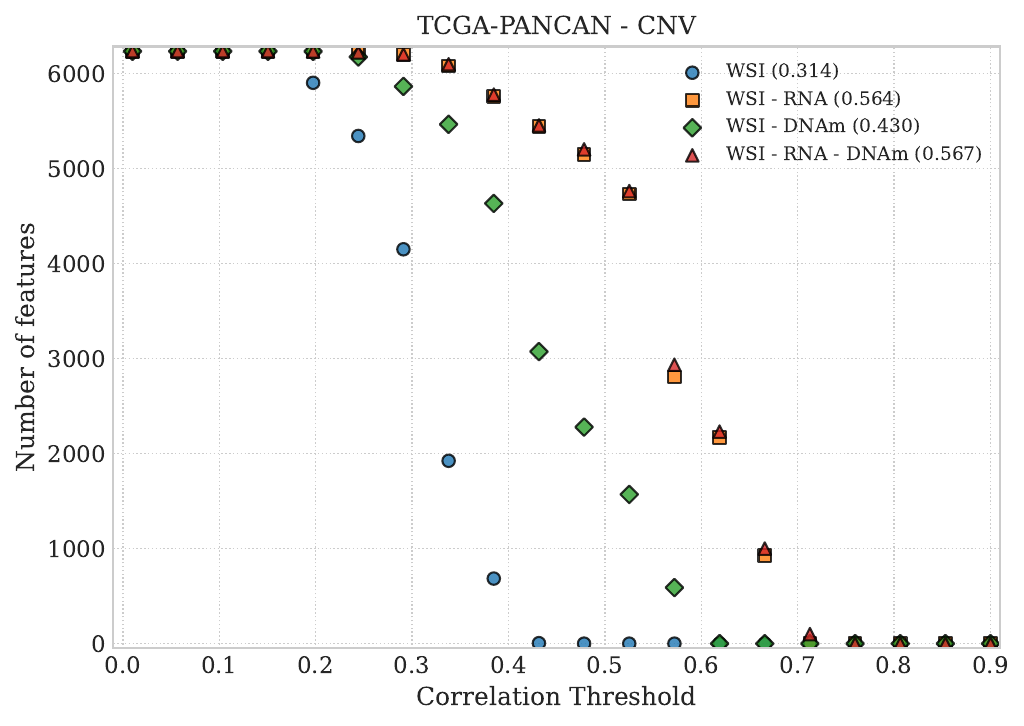}
\includegraphics[width=0.32\textwidth]{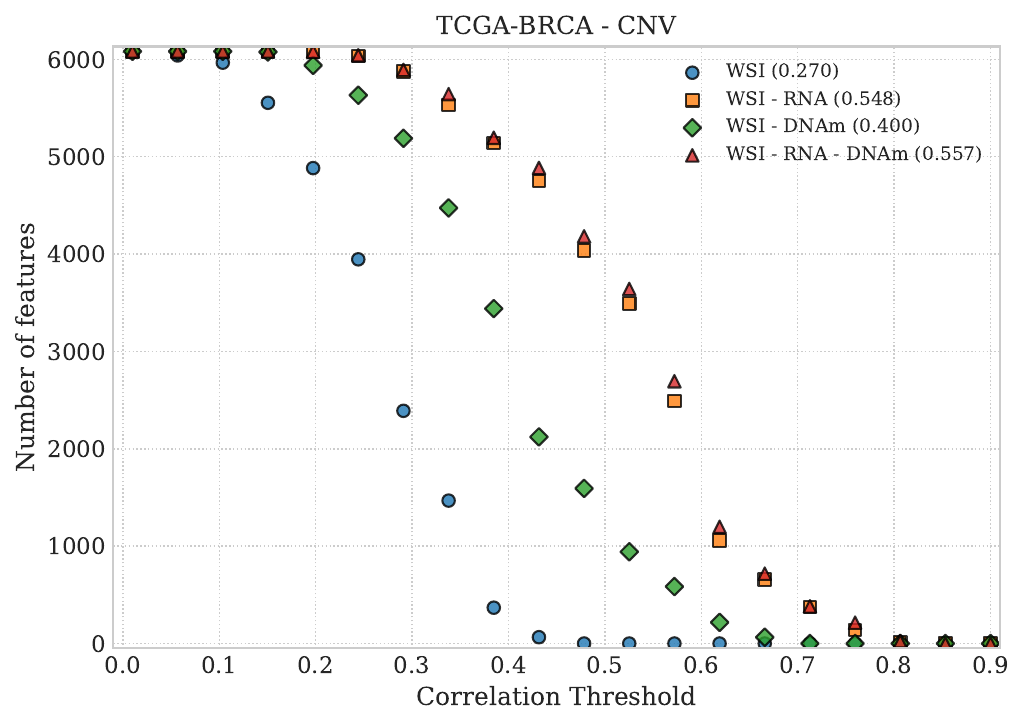}
\includegraphics[width=0.32\textwidth]{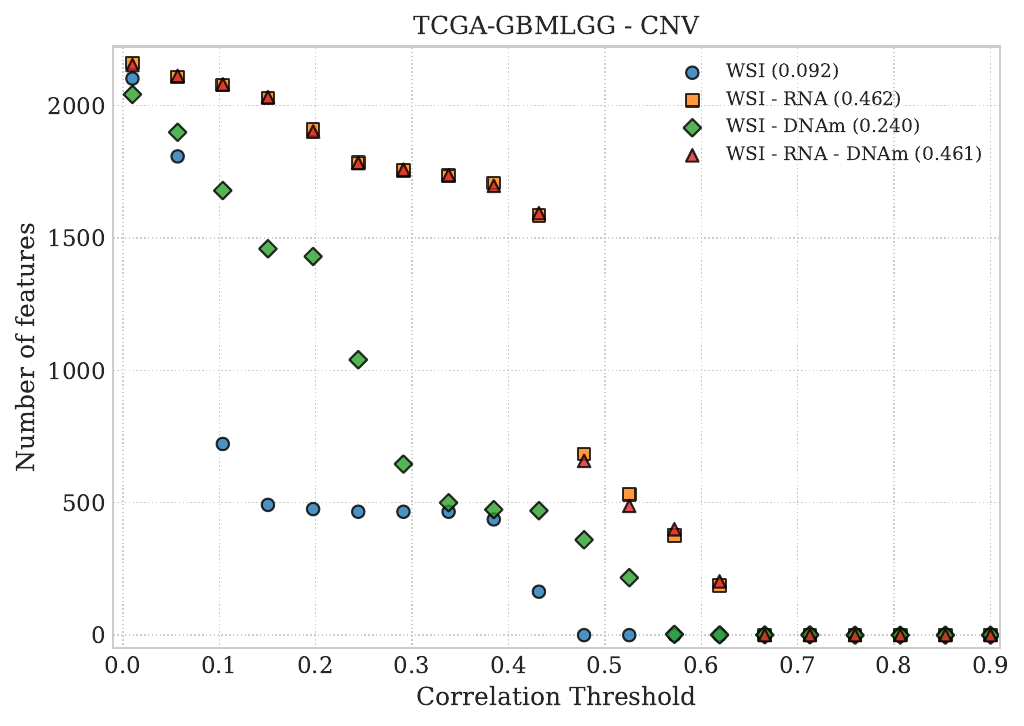}

\caption{\textbf{Reconstruction performance by modality and cohort.} 
Each plot shows the number of features (y-axis) whose Pearson correlation between predicted and true values exceeds the corresponding correlation threshold (x-axis).
Results are shown separately for RNA (top), DNAm (middle), and CNV (bottom), across different inputs combinations. 
Median correlation values are reported in the legend.}
\label{fig:recon_grid}
\end{figure*}
In contrast, by appropriately selecting masks, \morpheus\ enables flexible generation of arbitrary omics profile from WSI alone or conditioned on additional available omics modalities.
This supports a wide range of scenarios such as (WSI \(\rightarrow\) RNA, DNAm, CNV) or (WSI, RNA \(\rightarrow\) DNAm).
This design mirrors clinical reality, where data availability varies across patients and clinicians must cope with these constraints.
To quantitatively assess the relevance of the reconstructions, we first perform the same sanity check as proposed in \cite{hoang_prediction_2024}.
Specifically, we validate DNAm predictions by comparing hypermethylated and hypomethylated sites between real and reconstructed data across IDH subtypes. 
Results are provided in Appendix~\ref{app:san-check}.
We then evaluate reconstruction quality by measuring the Pearson correlation between predicted and true omics features across different input-modality combinations. 
In addition to the full cohort, denoted PANCAN, we report results on GBMLGG and BRCA, as cancer-specific subsets provide insight into performance under more homogeneous biological conditions.
Figure~\ref{fig:recon_grid} reports, for each modality, the number of reconstructed features whose correlation with the real data exceeds a given threshold.
While differences in datasets prevent direct quantitative comparison, the observed median correlation levels remain indicative of reconstruction fidelity.
For example, when reconstructing DNAm from WSI in the GBMLGG cohort, our model attains a median correlation of 0.459, which is comparable to values reported for this specific task \cite{hoang_prediction_2024}.
Moreover, the results show that incorporating additional modalities improves reconstruction quality, highlighting \morpheus’ ability to integrate complementary information through cross-modal interactions.
In particular, RNA and DNAm exhibit strong complementarity: when combined with WSI, each substantially improves the reconstruction of the other. 
In contrast, adding CNV to WSI yields smaller improvements.
When paired with another omics modality, CNV provides no additional benefit for reconstruction.
However, when CNV becomes the target, both RNA and DNAm contribute meaningfully, with RNA exerting the stronger influence.
Altogether, these results indicate that \morpheus\ learns robust cross-modal relationships, with multi-omics integration yielding synergistic effects.

\subsection{Cross-modal interactions}
\begin{figure*}[t]
\centering
\includegraphics[scale=0.23]{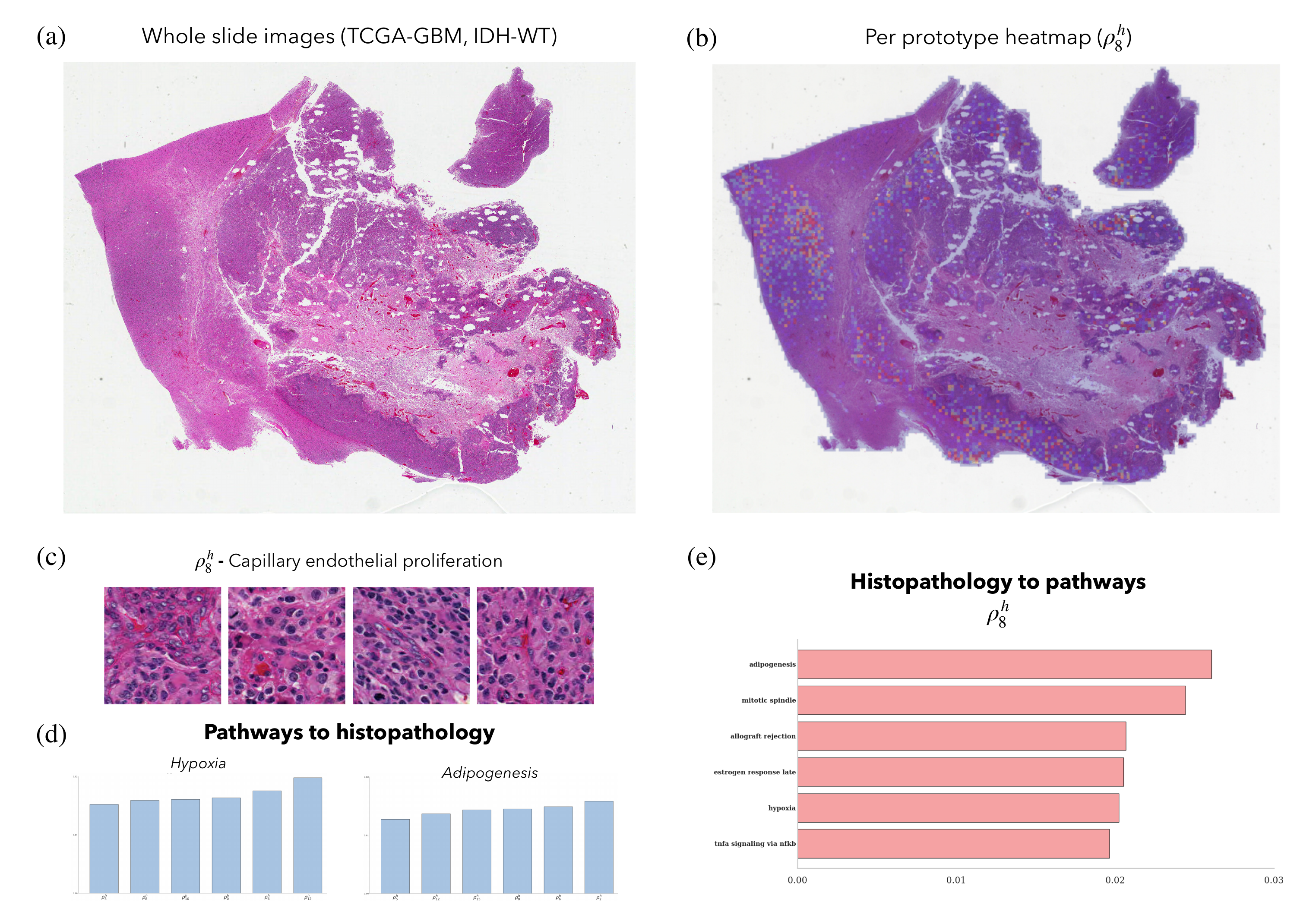}
\caption{\textbf{Cross-modal interaction visualization.} 
(a) WSI for a glioblastoma (IDH wild-type) patient. (b) Prototype heatmap for \(\rho_8^h\), associated with capillary endothelial proliferation. (c) Top-4 patches in the WSI corresponding to this prototype. (d) Top histopathology prototypes attending to hypoxia and adipogenesis. (e) Top RNA pathways attending to \(\rho_8^h\).}
\label{fig:cross-modal}
\end{figure*}

The flexibility of \morpheus\ enables a wide range of cross-modal analyses.
First, because histopathology patches are mapped to a compact set of latent prototypes by the Perceiver module, cross-attention weights can be directly visualized to generate prototype-level heatmaps (Figure~\ref{fig:cross-modal}b).
For each token, the most representative patches are identified via cross-attention weights (Figure~\ref{fig:cross-modal}c). 
Additional examples are provided in Appendix~\ref{app:proto}.

\morpheus\ also enables visualization of cross-modal interactions between omics and histopathology via self-attention weights.
For instance, we examine interactions from pathways to histopathology and from histopathology to pathways.
As an illustrative example, we consider prototype 8 (\(\rho_8^h\)), which captures capillary endothelial proliferation in a glioblastoma sample.
Inspecting attention weights reveals interactions with pathways related to hypoxia and adipogenesis (Figure~\ref{fig:cross-modal}e).
The association with hypoxia is well established in the glioblastoma literature, as it is recognized as a major prognostic factor \cite{cells6040045}.
While less frequently discussed, links to adipogenesis have also been reported \cite{tang_protein-based_2025}.
We can further visualize which histopathology tokens interact most strongly with these pathways (Figure~\ref{fig:cross-modal}d).
In particular, \(\rho_8^h\) consistently receives high attention from both hypoxia- and adipogenesis-related pathways.
Overall, \morpheus\ is designed to facilitate the analysis of biologically meaningful relationships across modalities, regardless of the number or type of inputs.

\section{Conclusion}
In this work, we introduce \morpheus, the first multimodal pre-training approach specifically tailored to cancer biology data.
Its core contribution is a masked omics modeling objective that enables the integration of histopathology with high-dimensional transcriptomic, methylomic, and genomic profiles.
By extending masked autoencoding principles to the molecular domain, \morpheus\ learns to reconstruct masked portions of omics data by leveraging information from available omics signals and the corresponding histopathology.
Following this multimodal pre-training, the shared encoder serves as a general-purpose backbone.
When fine-tuned, it outperforms strong baselines across a wide range of downstream tasks and modality combinations.
Beyond predictive performance, \morpheus\ also exhibits generative capabilities, enabling the reconstruction of any omics modality from histopathology, with or without additional omics inputs.
By effectively integrating heterogeneous data sources, this approach helps address clinical challenges related to data heterogeneity and scarcity.
Such flexibility supports more informed decision-making, which is essential in high-stakes medical applications.
Overall, these results highlight the promise of \morpheus\ for developing molecularly driven multimodal foundation models in oncology.

\clearpage

\bibliographystyle{unsrt}  
\bibliography{references}

\newpage
\appendix 
\section{Survival Analysis}
\label{app:survival}
\subsection{Generalities}
Survival analysis models the time until an event of interest occurs. 
This paradigm extends beyond medical research to fields such as social sciences, finance, and engineering, wherever time-to-event modeling is required.
Survival analysis shares similarities with regression in that it models an outcome based on features, but its defining characteristic is the handling of censored data. 
Censoring occurs when the event of interest is not observed within the study period. 
For example, in a clinical study tracking progression over two years, a patient who does not experience progression by the end of the study is considered right censored at two years. Censoring may also arise if a patient withdraws from the study or is lost to follow up. 
In such cases, the only information available is that the event did not occur up to a certain time point, known as the censoring time, beyond which the outcome remains unknown.
Hence, we aim to model the time-to-event \( T \in \mathbb{R}^+ \) in the presence of right-censoring.
Suppose the patient data is given by \( \mathcal{X} = \{x^g, x^h\} \). 
A survival dataset consists of triplets \( (T_i, \delta_i, \mathcal{X}_i) \), where \( T_i \) denotes the observed time and \( \delta_i \in \{0, 1\} \) indicates whether the event was observed (\( \delta_i = 1 \)) or censored (\( \delta_i = 0 \)).
The hazard function represents the instantaneous rate at which the event of interest occurs at time \( t \), given that it has not yet occurred. 
In continuous time, it is defined as:
\[
h(t \mid \mathcal{X}) = \lim_{\Delta t \to 0} \frac{P(t \leq T < t + \Delta t \mid T \geq t, \mathcal{X})}{\Delta t}.
\]
In a discrete-time setting, the hazard function simplifies to the conditional probability that the event occurs at time \( t \), given that it has not occurred before:
\[
h(t \mid \mathcal{X}) = P(T = t \mid T \geq t, \mathcal{X}).
\]
\subsection{Loss function}
We use a neural network \( \phi(\mathcal{X}_i) \) to estimate this hazard function based on the input data.
Thus, the time axis is discretized into \( Q \) non-overlapping intervals, typically denoted as \(( t_1, t_2, \dots, t_Q)\).
The neural network \( \phi(\mathcal{X}_i) \) outputs a vector of logits \( \mathbf{a}_i \in \mathbb{R}^Q \), where each component corresponds to a time interval. 
The hazard for interval \( q \) is modeled as:
\(
h_q(\mathcal{X}_i) = \sigma(a_{iq}),
\)
where \( \sigma(\cdot) \) denotes the sigmoid function.
The negative log-likelihood over the \( N \) samples, accounting for censoring \cite{9028113,gensheimer_scalable_2019}, is then given by:
\begin{align}
\ell_{hazard} ={}& - \sum_{i=1}^N \Bigg[
\delta_i \log\left(h_{q(i)}(\mathcal{X}_i)\right) + \sum_{j=1}^{q(i) - \delta_i} \log\left(1 - h_j(\mathcal{X}_i)\right)
\Bigg] \notag
\end{align}
where \(q(i)\) is the interval containing \(T_i\).
The first term models the likelihood of the event occurring in the observed interval for uncensored cases, while the second term accounts for survival up to (and including) the last known interval for both censored and uncensored cases. 
In \morpheus\, we adopt the PyCox \cite{kvamme_continuous_2021} implementation of this loss.

\subsection{Concordance Index}
The concordance index (C-index) is a commonly used metric to evaluate the predictive performance of survival models. 
It measures the model's ability to correctly rank survival times based on predicted risks. 
Intuitively, it reflects the proportion of all usable pairs of individuals for which the model's predictions are concordant with the actual order of observed events.
The C-index is then the proportion of concordant pairs among all comparable pairs.
A pair \( (i, j) \) is considered comparable if \( T_i < T_j \) and the event was observed for individual \( i \) (i.e., \( \delta_i = 1 \)); it is concordant if the predicted risk for \( i \) is higher than for \( j \), that is, if \( \hat{r}_i > \hat{r}_j \). 
Here, \( \hat{r}_i \) denotes the model's risk score for individual \( i \), such as the cumulative hazard or any scalar function indicating predicted risk.

\section{Cohorts description}
\label{app:cohorts}
\paragraph{Pre-training.}
Pre-training requires all modalities for each sample.
The corresponding data distribution is given in Table~\ref{tab:pre-train_distribution}.
\begin{table}[!h]
\centering
\scriptsize
\caption{\textbf{Distribution of pre-training patients across TCGA projects.}}
\label{tab:pre-train_distribution}
\begin{tabular}{l c}
\hline
\textbf{TCGA Project} & \textbf{Number of samples} \\ \hline
TCGA-BRCA & 594 \\
TCGA-THCA & 283 \\
TCGA-KIRC & 281 \\
TCGA-LGG  & 278 \\
TCGA-LUAD & 263 \\
TCGA-LUSC & 262 \\
TCGA-HNSC & 245 \\
TCGA-SKCM & 240 \\
TCGA-PRAD & 222 \\
TCGA-BLCA & 208 \\
TCGA-STAD & 191 \\
TCGA-LIHC & 188 \\
TCGA-COAD & 155 \\
TCGA-KIRP & 153 \\
TCGA-CESC & 148 \\
TCGA-SARC & 135 \\
TCGA-PAAD & 99  \\
TCGA-PCPG & 89  \\
TCGA-UCEC & 88  \\
TCGA-TGCT & 85  \\
TCGA-ESCA & 84  \\
TCGA-THYM & 68  \\
TCGA-READ & 50  \\
TCGA-UVM  & 48  \\
TCGA-MESO & 41  \\
TCGA-OV   & 41  \\
TCGA-KICH & 37  \\
TCGA-GBM  & 34  \\
TCGA-ACC  & 31  \\
TCGA-UCS  & 30  \\
TCGA-DLBC & 26  \\
TCGA-CHOL & 21  \\
\hline
\end{tabular}
\end{table}
\paragraph{Biomarker prediction.}
Cohorts detailed in Table~\ref{tab:few_shot_subtyping} consist of patients excluded from the pre-training and for whom the WSI and the corresponding biomarker are available. 
\begin{table}[!h]
\centering
\small
\caption{\textbf{Cohort sizes for biomarker prediction tasks.}}
\label{tab:few_shot_subtyping}
\begin{tabular}{c l}
\hline
\textbf{Biomarker} & \textbf{Number of samples} \\ \hline
\textbf{Brain} \text{\scriptsize (IDH)}   & 324 (WT = 159 \& Mutant = 165) \\
\textbf{Brain} \text{\scriptsize (ATRX)}    & 268 (WT = 187 \& Mutant = 81) \\
\textbf{Brain} \text{\scriptsize (1p19q)}  & 349 (non-codel = 292 \& codel = 57) \\
\textbf{Lung} & 378 (LUAD = 191 \& LUSC = 187) \\
\textbf{Breast} & 351 (IDC = 281 \& ILC = 70) \\
\hline
\end{tabular}
\end{table}
\paragraph{Survival analysis.}
For survival analysis and omics reconstruction, we retain only samples with all modalities available to ensure a fair comparison across configurations.
\begin{table}[!h]
\centering
\small
\caption{\textbf{Cohort sizes for survival analysis.}}
\label{tab:survival_cohorts}
\begin{tabular}{l c}
\hline
\textbf{TCGA Project} & \textbf{Number of samples} \\ \hline
TCGA-BRCA & 348 \\
TCGA-GBMLGG   & 223 \\
TCGA-COADREAD &  115 \\
TCGA-STAD & 130 \\
TCGA-LUAD & 165 \\
\hline
\end{tabular}
\end{table}

\paragraph{Omics reconstructions.} 
We keep every sample not in the pre-training set that contains the four modalities, yielding the distribution outlined in Table~\ref{tab:reconstruction_samples}.
To ensure robustness, only DNA methylation features with standard deviation greater than 0.15 and CNV features with standard deviation greater than 0.1 are retained, which explains the cohort-specific differences in feature counts.
\begin{table}[!h]
\centering
\caption{\textbf{Cohort sizes for reconstruction experiments.}}
\label{tab:reconstruction_samples}
\begin{tabular}{l c}
\hline
\textbf{TCGA Project} & \textbf{Number of Patients} \\ \hline
TCGA-GBMLGG   & 228 \\
TCGA-BRCA     & 408 \\
TCGA-PANCAN   & 3,302 \\
\hline
\end{tabular}
\end{table}

\section{Omics preprocessing}
\label{app:preprocessing}

Hereafter we give in-depth information about how modalities are preprocessed.

\paragraph{Gene expression (RNA).} Bulk RNA profiles for all TCGA cohorts were retrieved from the UCSC Xena repository \cite{goldman_visualizing_2020}. 
Expression values were measured on the Illumina HiSeq 2000 RNA Sequencing platform and underwent \( log_2(x + 1) \) RSEM normalization \cite{li_rsem_2011}. 
To capture pathway level signals, genes were aggregated into \(N_g = 50\) Hallmark pathways \cite{msigdb}.

\paragraph{DNA methylation (DNAm).} Building on prior work \cite{vale-silva_long-term_2021,hoang_prediction_2024}, we confined DNA methylation \(\beta\)-values to the 24 655 CpG probes shared by the Illumina Infinium 27K, 450K, and EPIC (850K) arrays, discarding any probe with over 99\% missing data. 
From this set, we selected the 7 150 most variable probes and excluded those on chromosomes X and Y. 
The remaining probes were then grouped by chromosomal position into \(N_m = 50 \) clusters of roughly equal size (see Table \ref{tab:dnam_chromosome_clusters}). 
Because \(\beta\)-values naturally lie in [0, 1], no further normalization was required.
\begin{table}[!h]
\centering
\scriptsize
\caption{\textbf{Number of clusters per chromosome}}
\label{tab:dnam_chromosome_clusters}
\begin{tabular}{l | c}
\hline
\textbf{Chromosome} & \textbf{\# Groups} \\ \hline
chr1  & 6 \\
chr2  & 3 \\
chr3  & 3 \\
chr4  & 2 \\
chr5  & 2 \\
chr6  & 3 \\
chr7  & 2 \\
chr8  & 2 \\
chr9  & 2 \\
chr10 & 2 \\
chr11 & 3 \\
chr12 & 3 \\
chr13 & 1 \\
chr14 & 1 \\
chr15 & 1 \\
chr16 & 2 \\
chr17 & 3 \\
chr18 & 1 \\
chr19 & 4 \\
chr20 & 2 \\
chr21 & 1 \\
chr22 & 1 \\
\hline
\end{tabular}
\end{table}
\paragraph{Copy Number Variations (CNV).} 
Analogous to our DNA methylation pipeline, we process CNV data by first selecting the 6,750 genes exhibiting the highest variance across the pre-training cohort. 
We then organize these genes by chromosomal position into \(N_c = 50\) clusters of roughly equal size, ensuring balanced representation from each chromosome (see Table \ref{tab:cnv_chromosome_clusters} for details). 
Missing CNV values were imputed to the normal diploid value (2), and all copy‐number ratios were transformed as \(log_{10}(\text{CNV}/2 + 1)\).
\begin{table}[!h]
\centering
\scriptsize
\caption{\textbf{Number of CNV gene clusters per chromosome or chromosome groups}}
\label{tab:cnv_chromosome_clusters}
\begin{tabular}{l | c}
\hline
\textbf{Chromosome(s)}           & \textbf{\# Groups} \\
\hline
chr1                    & 8           \\
chr7                    & 7           \\
chr3                    & 4           \\
chr17                   & 5           \\
chr8                    & 5           \\
chr20                   & 4           \\
chr12                   & 3           \\
chr19                   & 3           \\
chr6                    & 3           \\
chr11                   & 2           \\
chr4; chr5              & 1           \\
chr2; chr22; chr21      & 1           \\
chr14; chr15; chr16     & 1           \\
chr9; chr10             & 1           \\
chr13                   & 1           \\
chr18                   & 1           \\
\hline
\end{tabular}
\end{table}
\section{Architecture details}
\label{app:tangle}
\subsection{\(\text{TANGLE}^{\dagger}\) details}
The original \textsc{Tangle} \cite{jaume_transcriptomics-guided_2024} uses a task-specific patch encoder and proprietary in-house data, making direct comparison difficult. 
For fairness, we rely on the official public implementation\footnote{\url{https://github.com/mahmoodlab/TANGLE}} and retrain \textsc{Tangle} with UNIv2 \cite{chen_towards_2024} embeddings, using the same data split and hyperparameters as \morpheus.
Contrastive loss temperature and other \textsc{Tangle}-specific parameters follow the best settings reported in \cite{jaume_transcriptomics-guided_2024}, and the entire network is fine-tuned end-to-end.
\subsection{Neural network architectures baselines with all modalities}
We use a simple MLP for each omics modality, composed of two linear layers (input → 128 and 128 → \(d\)), each followed by LayerNorm, ReLU, and dropout.
For WSI, we mean-pool the patch embeddings and project the resulting vector into the \(d\)-dimensional latent space using a linear layer.

\clearpage
\section{Hyperparameters}
\label{app:hparams}
\begin{table}[!h]
\small
\centering
\caption{\textbf{Hyperparameters for \morpheus\ pre-training.}}
\label{tab:hparams_pretrain}
\begin{tabular}{l |c}
\hline
\textbf{Hyperparameter} & \textbf{Value} \\ \hline
\(N_h\)                      & 16 \\
\(N_g\)                & 50 (Hallmarks) \\
\(N_m\)               & 50 \\
\(N_c\)         & 50 \\
\(f_{\text{enc}}\) & UNIv2 \cite{chen_towards_2024} \\
\hline
Epochs                       & 200 \\
Masking ratio \(r\)               & 0.75 \\
Concentration parameter \(\alpha\) & 1 \\
Batch size                       & 128 \\
Num patches               & 1024 \\
\hline
Layers            & 1 \\
Attention heads        & 8 \\
Multi-layer perceptron dimension & 256 \\
Embedding dimension \(d\) & 256 \\
Dropout rate & 0.15 \\
\hline
Weight decay           & 1e-3 \\
Optimizer        & AdamW \\
Warmup epochs & 10 \\
Learning rate schedule & Cosine \\
Learning rate (start) & 5e-5 \\
Learning rate (post warmup) & 5e-4 \\
Learning rate (final) & 1.5e-4 \\ \hline
Training time (in hours) & \(\approx\) 18  \\ \hline
\end{tabular}
\end{table}

\begin{table}[!h]
\centering
\small
\caption{\textbf{Hyperparameters for biomarker prediction.}}
\label{tab:hparams_clf}
\begin{tabular}{l | c}
\hline
\textbf{Hyperparameter} & \textbf{Value} \\
\hline
Epochs                       & 5 \\
Batch size                       & 32 \\
Num patches               & 1024 \\
Dropout rate & 0.35 \\
\hline
Weight decay           & 1e-2 \\
Optimizer        & AdamW \\
Learning rate & 5e-5 \\ \hline
\end{tabular}
\end{table}

\begin{table}[!h]
\centering
\small
\caption{\textbf{Hyperparameters for few-shot biomarker prediction.}}
\label{tab:hparams_few_shot}
\begin{tabular}{l | c}
\hline
\textbf{Hyperparameter} & \textbf{Value} \\
\hline
Epochs                       & 5 \\
Batch size                       & 1 \\
Num patches               & 1024 \\
Dropout rate & 0.35 \\
\hline
Weight decay           & 1e-2 \\
Optimizer        & AdamW \\
Learning rate & 5e-5 \\ \hline
\end{tabular}
\end{table}
\begin{table}[!h]
\centering
\small
\caption{\textbf{Hyperparameters for survival analysis.}}
\label{tab:hparams_surv}
\begin{tabular}{l | c}
\hline
\textbf{Hyperparameter} & \textbf{Value} \\
\hline
Epochs                       & 20 \\
Batch size                       & 32 \\
Num patches               & 1024 \\
Dropout rate & 0.35 \\
Num intervals & 5 \\
\hline
Weight decay           & 1e-2 \\
Optimizer        & AdamW \\
Warmup epochs & 5 \\
Learning rate schedule & Cosine \\
Learning rate (start) & 1e-5 \\
Learning rate (post warmup) & 5e-5 \\
Learning rate (final) & 6e-6 \\ \hline
\end{tabular}
\end{table}
\section{Sensitivity analysis}
\label{app:sensitivity}
\begin{figure}[!h]
\centering
\includegraphics[scale=0.42]{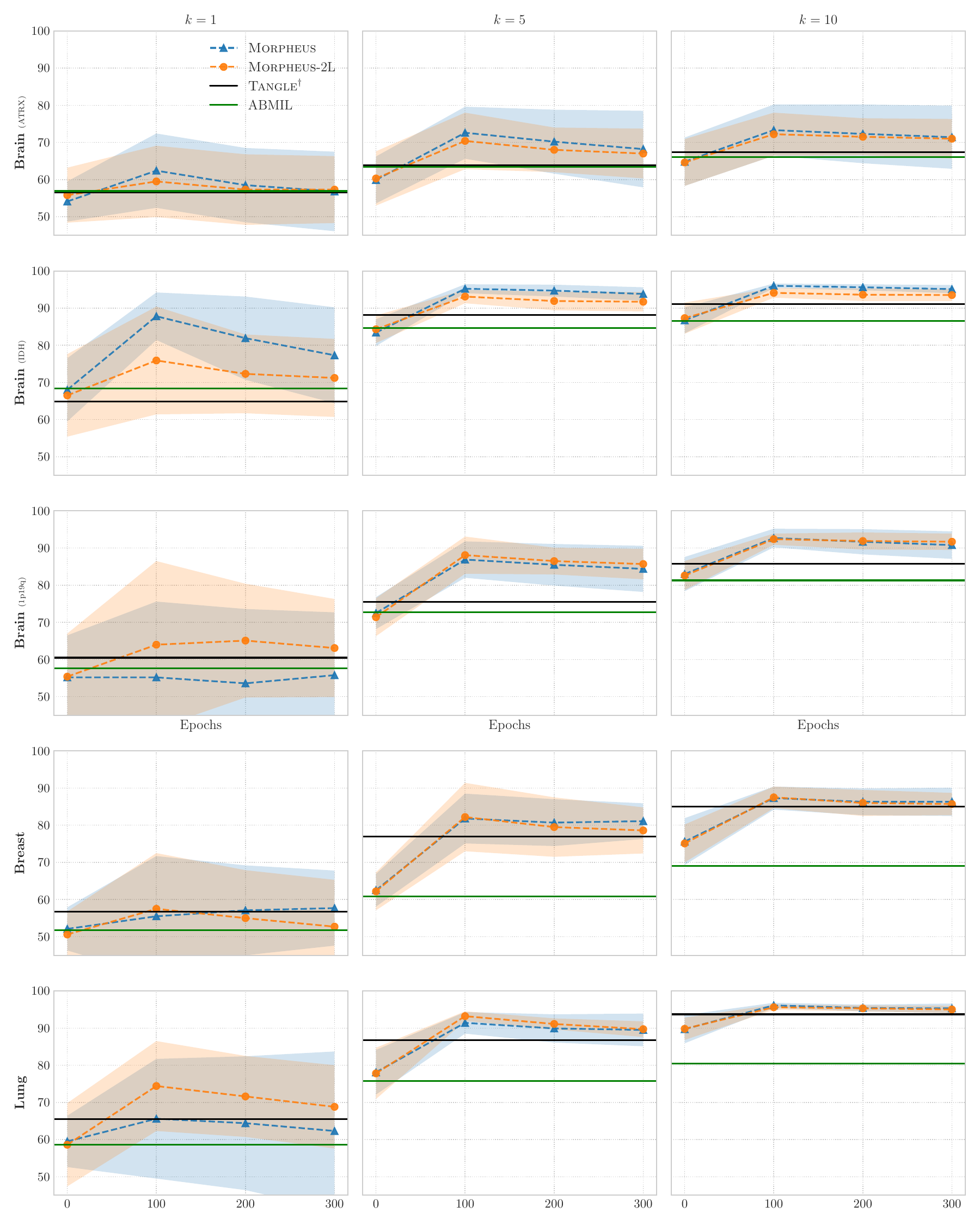}
\caption{\textbf{Sensitivity analysis on the number of pre-training epochs.} 
Few-shot biomarker prediction performance with increasing number of epochs, compared against ABMIL (supervised) \cite{ilse_attention-based_2018} and \textsc{Tangle}$^\dagger$ (pre-trained) \cite{jaume_transcriptomics-guided_2024} baselines.
}
\label{fig:sensitivity}
\end{figure}
Figure~\ref{fig:sensitivity} reports few-shot performance for pre-training durations ranging from 0 (no pre-training) to 300 epochs.
We observe that longer pre-training does not consistently improve AUC and can even degrade performance in the extreme one-shot setting (\(k = 1\)) for IDH prediction.
Except for 1p19q, breast, and lung subtyping at (\(k=1\)), both \morpheus\ variants perform better than all baselines for every pre-training duration.
When trained from scratch, AUC falls back to the level of conventional supervised baselines, highlighting the substantial gains provided by the multimodal pre-training step.

\section{Sanity Check for Methylation Prediction}
\label{app:san-check}

We follow the same sanity check procedure as in DEPLOY \cite{hoang_prediction_2024}.
For DNA methylation prediction in brain tumors, we assess whether the model captures the expected differential hypermethylation patterns between IDH-mutant (\(n=159\)) and IDH-WT (\(n=66\)) gliomas. 
Specifically, we identify CpG sites that are significantly hyper- or hypomethylated in the real data and evaluate whether the predicted methylation values reflect the same direction of change. This analysis is repeated across all modality combinations, with results summarized in Table~\ref{tab:methylation_sanity_check}.

\begin{table}[h]
\caption{\textbf{Differential methylation change.} Accuracy (\%) in predicting whether CpG sites are hyper- or hypomethylated across different modality combinations.}
\label{tab:methylation_sanity_check}
\centering
\small
\begin{tabular}{l c}
\hline
\textbf{Modality Combination} & \textbf{Accuracy (\%)} \\ \hline
WSI                      & 97.6 \\
WSI + RNA               & 98.2 \\
WSI + CNV                & 97.3 \\
WSI + RNA + CNV          & 98.1 \\
\hline
\end{tabular}
\end{table}
\clearpage
\section{Top patches per prototype}
\label{app:proto}

\begin{figure}[ht]
\centering
\includegraphics[scale=0.28]{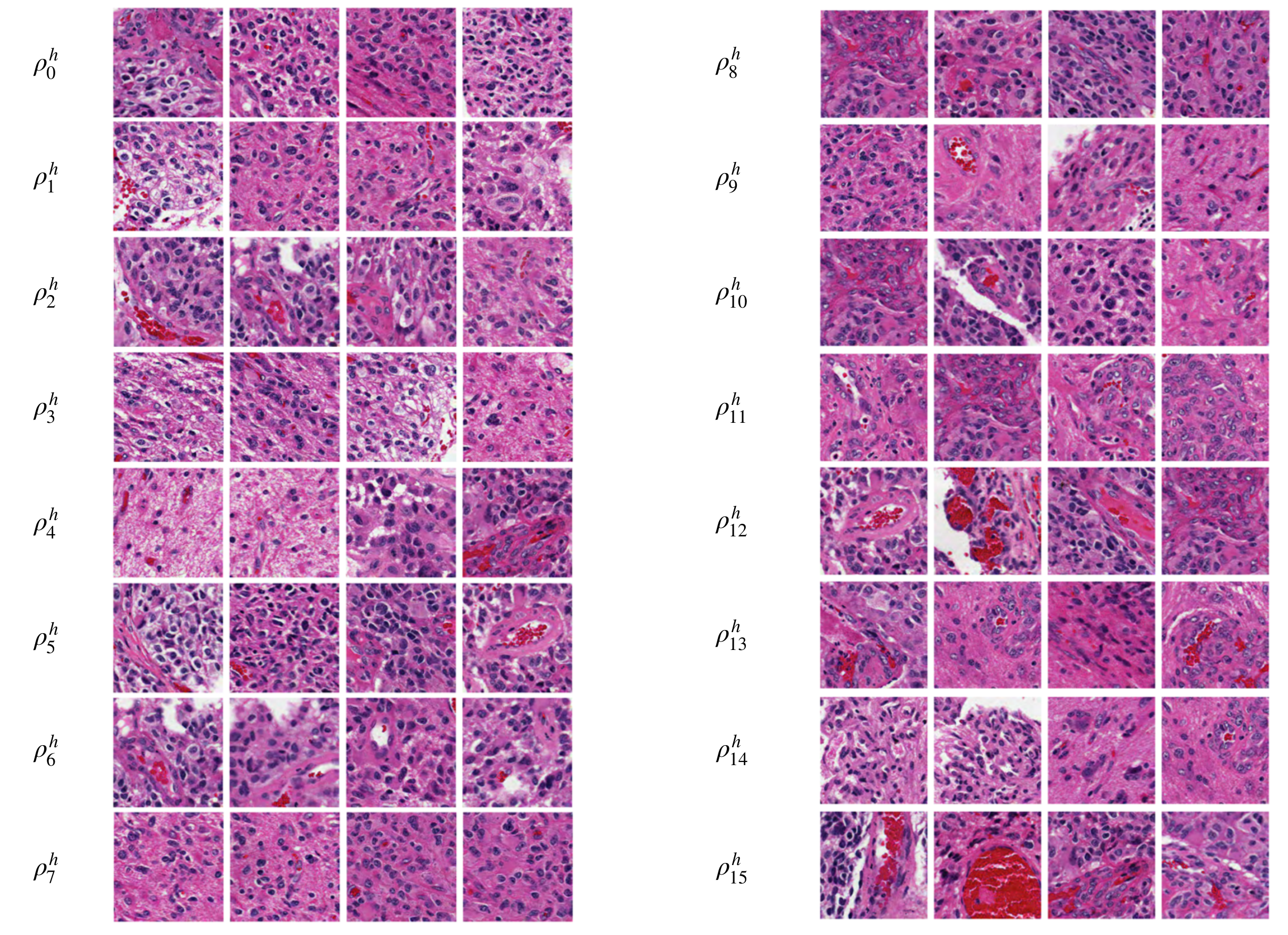}
\caption{\textbf{Prototype-specific top patches visualizations.} 
For each histopathology prototype, we show the four WSI patches with the highest cross-attention scores for a glioblastoma patient, illustrating the visual patterns most strongly associated with each prototype.
}
\label{fig:prototypes}
\end{figure}

\end{document}